\documentclass[journal]{IEEEtran}
\usepackage{bm}
\usepackage{multirow}
\usepackage{color}
\usepackage{graphicx}
\usepackage{amssymb}
\usepackage{amsmath}
\usepackage{url}

\ifCLASSINFOpdf
\else
\fi

\begin{document}
\title{MFHI: Taking Modality-free Human Identification as Zero-shot Learning}

\author{Zhizhe Liu,
        Xingxing Zhang,
        Zhenfeng Zhu$^*$,
        Shuai Zheng,
        Yao Zhao,~\IEEEmembership{Senior Member,~IEEE,}
        and~Jian~Cheng% <-this % stops a space
\thanks{This work was supported in part by Science and Technology Innovation 2030 – New Generation Artificial Intelligence Major Project under Grant 2018AAA0102101, in part by the National Natural Science Foundation of China under Grant No. 61976018.}    
\thanks{Z. Liu, Z. Zhu, S. Zheng and Y. Zhao are with the Institute of Information Science, Beijing Jiaotong University, Beijing 100044, China, and also with the Beijing Key Laboratory of Advanced Information Science and Network Technology, Beijing Jiaotong University, Beijing 100044, China (E-mail: zhzliu@bjtu.edu.cn; zhfzhu@bjtu.edu.cn; zs1997@bjtu.edu.cn; yzhao@bjtu.edu.cn). (\emph {Corresponding author: Zhenfeng Zhu.})}% <-this % stops a space
\thanks{X. Zhang is with the Department of Computer Science and Technology, Tsinghua University, Beijing 100084, China. E-mail: xxzhang2020@mail.tsinghua.edu.cn.}
\thanks{Jian Cheng is with the National Laboratory of Pattern Recognition, Institute of Automation, Chinese Academy of Sciences, Beijing, China (E-mail: jcheng@nlpr.ia.ac.cn).}}% <-this % stops a space


\markboth{Journal of \LaTeX\ Class Files,~Vol.~14, No.~8, August~2015}%
{Shell \MakeLowercase{\textit{et al.}}: Bare Demo of IEEEtran.cls for IEEE Journals}



\maketitle

%\begin{abstract}
%	Existing face identification methods predominantly classify a queried face \underline{image} to a specific identity in an \underline{image} gallery set.
%	This is seriously limited for the scenario where only a \underline{textual description} of the query or an attribute gallery set is available in a wide range of video surveillance applications.
%	However, very few efforts have been devoted towards this direction.
%	In this work, we take an initial attempt, and formulate \textbf{M}odality-\textbf{F}ree \textbf{F}ace \textbf{I}dentification (named MFHI) as a generic zero-shot learning model without data scale limitation.
%	Meanwhile, it is capable of bridging the visual and semantic modalities by identity prototype learning.
%	In addition, the semantics-guided spatial attention is enforced to obtain representations with both high global category-level and local attribute-level discrimination.
%	We design and conduct an extensive group of experiments on \textit{CelebA} and \textit{LFWA}, demonstrating that our method outperforms a wide variety of state-of-the-art methods on modality-free face identification task.
%	Moreover, we extend MFHI to person re-identification (re-ID) task which also exists the same limitations.
%	Comprehensive evaluation experiments prove that our MFHI posseses significant generalization ability. 
%	Code is available in \url{https://github.com/1111122222a/MFHI}.
%\end{abstract}

\begin{abstract}
	Human identification is an important topic in event detection, person tracking, and public security.
	There have been numerous methods proposed for human identification, such as face identification, person re-identification, and gait identification.
	Typically, existing methods predominantly classify a queried \underline{image} to a specific identity in an \underline{image} gallery set (\textit{I2I}).
	This is seriously limited for the scenario where only a \underline{textual} description of the query or an \underline{attribute} gallery set is available in a wide range of video surveillance applications (\textit{A2I} or \textit{I2A}). 
	However, very few efforts have been devoted towards modality-free identification, i.e., identifying a query in a gallery set in a scalable way.
	In this work, we take an initial attempt, and formulate such a novel \textbf{M}odality-\textbf{F}ree \textbf{H}uman \textbf{I}dentification (named MFHI) task as a generic zero-shot learning model in a scalable way.
	Meanwhile, it is capable of bridging the visual and semantic modalities by learning a discriminative prototype of each identity.
	In addition, the semantics-guided spatial attention is enforced on visual modality to obtain interpretable representations with both high global category-level and local attribute-level discrimination.
	Finally, we design and conduct an extensive group of experiments on two common challenging identification tasks, including face identification and person re-identification, demonstrating that our method outperforms a wide variety of state-of-the-art methods on modality-free human identification.
	%Extensive experiments on both face identification and re-ID tasks verify that our MFHI is a win-win formulation, i.e, not only can the limitation of modality be removed by MFHI with a serious difference from conventional methods, but the improvements 
	%Moreover, we extend MFHI to person re-identification (re-ID) task which also exists the same limitations.
	%Comprehensive evaluation experiments prove that our MFHI posseses significant generalization ability. 
	%	Code is available in \url{https://github.com/1111122222a/MFHI}.
\end{abstract}


% \fi
\IEEEpeerreviewmaketitle

\section{Introduction}\label{Introduction}

\IEEEPARstart{G}{enerally}, human identification aims to verify the identity of a person based on one or more biometric features, e.g., face~\cite{deng2019arcface}, gait~\cite{6737218}, and person image~\cite{zeng2020hierarchical}.
It has been widely used in various areas such as stations, schools, and companies, since human identification is crucially important in surveillance~\cite{quach2021dyglip}, activity analysis~\cite{Chen2010Time}, and person search~\cite{zeng2020hierarchical}.
%--------------------Figure 1---------------------------%
\begin{figure}[!htbp]
	\begin{center}
		\includegraphics[width=3.1in,height=3.0in ]{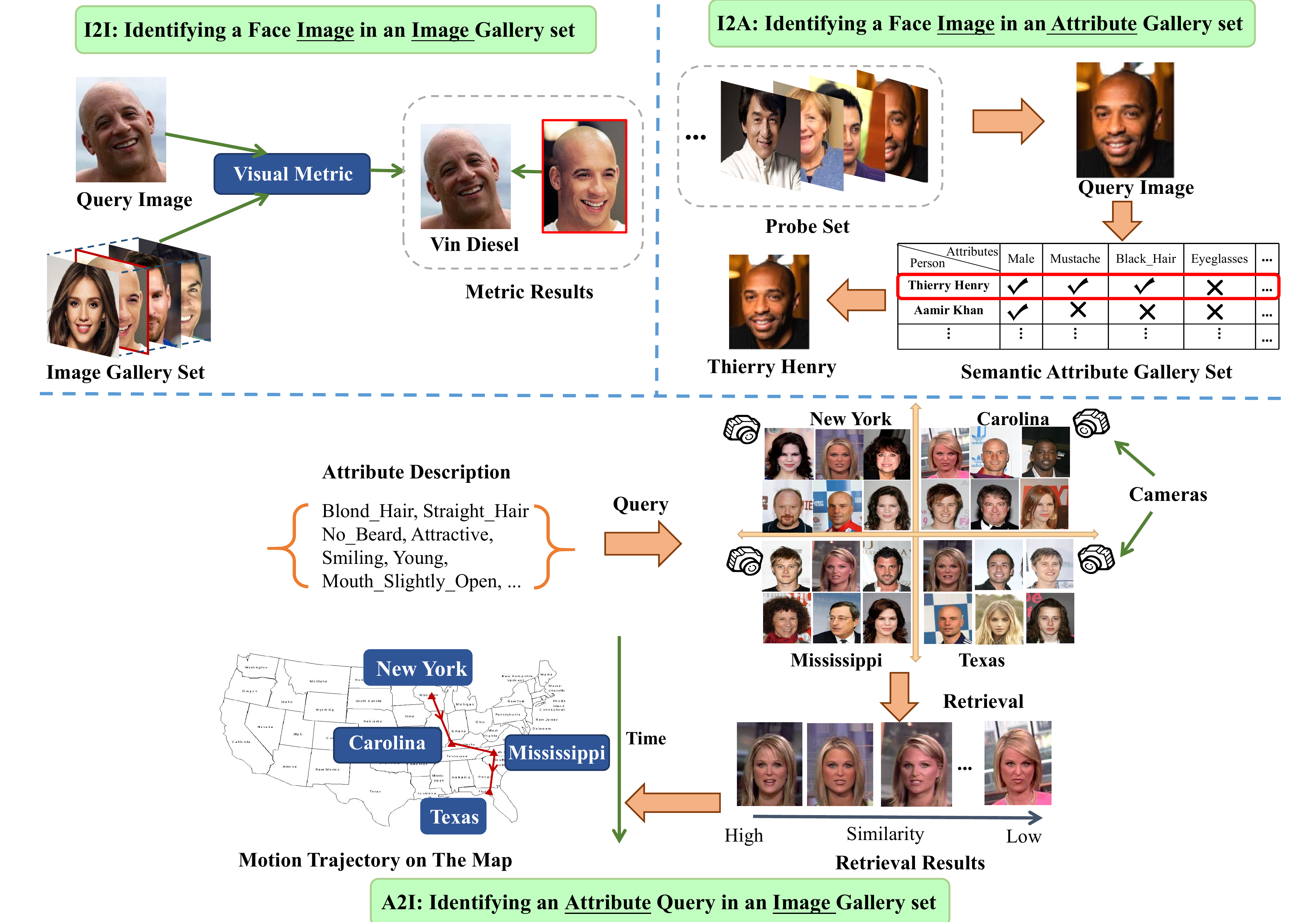}
	\end{center} %\linewidth
	\vskip -0.15in
	\caption{The illustration of modality-free face identification, including I2I, I2A, and A2I three scenarios.}
	\label{fig:MFHI}
	\vskip -0.15in
\end{figure}
%--------------------Figure 1---------------------------%
Especially, as two typical representatives of human identification, face identification and person re-identification (re-ID) play a key role in such an important topic, and have received extensive attention and success recently.
In this work, we focus on these two challenging identification tasks.

In recent years, owing to significant advances in deep learning and discriminative learning approaches~\cite{deng2019arcface}, convolutional neural networks (CNN) have increased the face identification to an unprecedented level.
Generally, a face identification method involves a training set, a gallery set, and a probe set, where the gallery set of target identity must be collected before classifying a query from a probe set.
There are open-set and close-set two settings in face identification task.
For the close-set, the queried identity must appear in the training set, while for the open-set, it generally never appears. This is very important and popular since collecting sufficient training data for all possible identities would be very difficult.  
Obviously, face identification in open-set is more suitable for complex scenarios in daily life, and has been widely explored~\cite{deng2019arcface}.
Thus, we mainly study open-set case in this work.

%--------------------Figure 2---------------------------%
\begin{figure}[htbp]
	\vskip -0.15in
	\begin{center}
		\includegraphics[width=0.5\textwidth,height=0.3\textwidth]{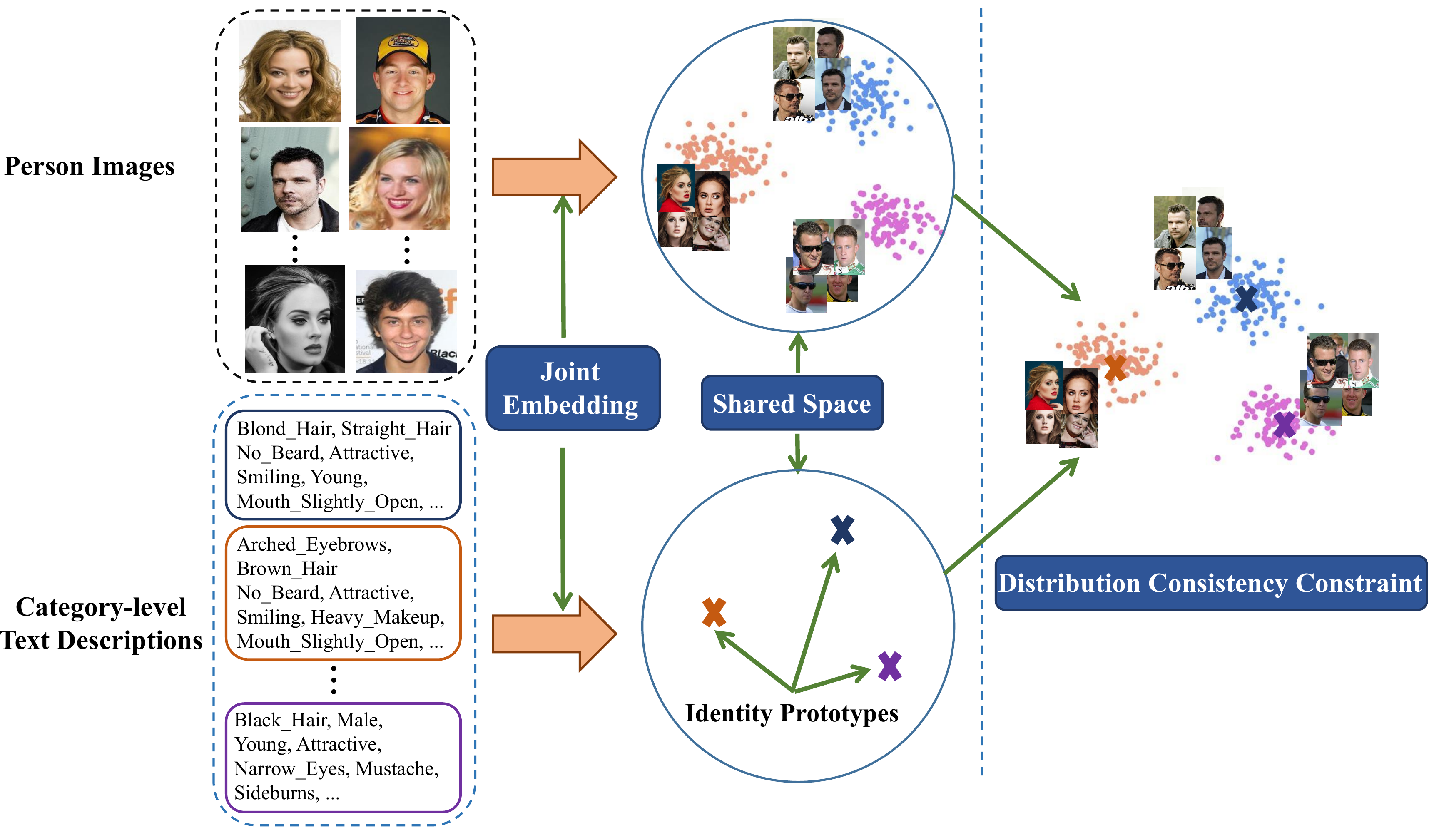}
	\end{center} %\linewidth
	\vskip -0.15in
	\caption{The illustration of joint embedding learning in modality-free identification. For example, images and texts are embedded into a shared space, and the distributions of two modalities are consistent in the shared space.}
	\vskip -0.08in
	\label{fig:shared_space}
\end{figure}
%--------------------Figure 2---------------------------%

Similar to face identification, re-ID also obtains significant improvements with the help of rich deep visual features, and has been widely used in many situations such as long-term multi-camera tracking~\cite{quach2021dyglip} and activity analysis~\cite{Chen2010Time}.
Recently, due to the increasing number of surveillance cameras, a large amount of raw video data is continually accumulated every day.
Thus, re-ID becomes more challenging and essential in real-world applications.
Most existing re-ID methods focus on image queries (probe set)~\cite{zeng2020hierarchical}, and aim to retrieve the images with the same identity to the queried image from the gallery set.
Generally, given a queried image, these methods calculate pairwise visual similarity scores between the queried image and every gallery image in the gallery set, and then treat the gallery images with the higher similarity scores as possible matches.
Specially, it is worth noticing that the person identities in training and gallery sets are also disjoint in re-ID. 

%Likewise, a number of methods have been proposed for open-set face identification~\cite{roychowdhury2015face,liu2017sphereface,wang2018cosface,deng2019arcface}.
However, existing face identification and re-ID tasks predominantly focus on how to accurately classify a queried image to a specific identity in an image gallery set.
Actually, in many practical scenarios, only a textual description of the query, instead of an image, is available. Likewise, only a gallery set that collects all textual descriptions of target identities, but not images, is provided.
This is closely related to a wide range of video surveillance applications.
Taking face identification as an example, as shown in Fig.~\ref{fig:MFHI}, based on an pre-collected \underline{image} gallery set, the first case aims to verify  the identity of the query \underline{image}.
%Taking face identification as an example, as shown in Fig.~\ref{fig:MFHI}, the first case is based on a pre-collected gallery set to verify the identity of the query image, and aims to classify a queried \underline{image} to a specific identity in an \underline{image} gallery set.
The goal of the second case is to classify a queried face \underline{image} to a specific identity in an \underline{attribute} gallery set, since the corresponding image gallery set (i.e., visual information) cannot been provided for privacy protection.
The third case is to retrieve images with the same identity from an \underline{image} gallery set, if a queried face \underline{attribute} description is given.
%\textcolor{red}{The same various complex scenarios also appear in re-ID.}
%the third case is to classify a queried face \underline{attribute} to a specific identity in an \underline{image} gallery set.
We name face identification in all these scenarios as \textbf{modality-free face identification}.
For example, given a list of textual descriptions from witnesses, i.e., criminal portrait, we aim to perform face identification from a provided image gallery set, and further obtain the motion trajectory of criminal on the map.
In fact, the current human identification methods mainly focus on the independent Image$\rightarrow$Image and Attribute$\rightarrow$Image identification and have obtained impressive results. However, such independent models obviously won’t allow flexible transfer between different recognition tasks such as Image$\rightarrow$Image and Attribute$\rightarrow$Image. Meanwhile, in realistic identification scenarios, we usually require the model with strong robustness and generalizability to address various identification tasks. From a practical point of view, it means that the ability to transfer between tasks is necessary when developing a flexible human identification model.
%Unfortunately, existing face identification methods fail to address such a modality-free challenge, since they often just consider an image modality while enhancing the discriminability of learned face features.

Furthermore, considering that only a small ration of identities can be used for model training in open-set setting, we take an initial attempt, and formulate such a modality-free human identification task as a zero-shot learning (ZSL) model. Generally, ZSL~\cite{zhang2017learning} aims to classify objects which may not have any training samples. Although ZSL has obtained great progress~\cite{Xie_2019_CVPR}, existing ZSL methods are suboptimal for our problem.
\textit{First}, conventional ZSL setting usually assigns a label for a queried image within a small scale categories, while human identification is actually a more challenging and larger scale ZSL problem.
\textit{Second}, existing ZSL setting is with both low inter-class similarity and small intra-class variation, while it is contrary for human identification problem.
This results in the image features learned by existing ZSL methods would be indiscriminative~\cite{zhang2017learning,annadani2018preserving} for such a challenging identification task. 
\textit{Third}, existing ZSL methods only consider classifying a queried image to a category in an attribute gallery set~\cite{sung2018learning}, i.e., the second case in Fig.~\ref{fig:MFHI}. They cannot be directly extended for modality-free human identification.

Motivated by the above observations, in this work, we aim to formulate such a modality-free human identification task as a \textbf{generic} ZSL model.
Concretely, inspired by ArcFace for face identification in~\cite{deng2019arcface}, we also introduce an additive angular margin in our model for high global category-level discrimination in large-scale category scenarios. It actually has a clear geometric interpretation due to its exact correspondence to geodesic distance on a hypersphere.
Moreover, to maximize the human identity separability, the local attribute-level discrimination is additionally considered in MFHI by learning an attribute-driven spatial attention. It is capable of capturing the distribution inconsistence between identities, thus enhancing the discriminative power of visual features.
Finally, for modality-free human identification, we innovatively bridge the visual and semantic modalities, i.e., images and texts, by learning a shared space as shown in Fig.~\ref{fig:shared_space}.
It should be noticed that some previous cross-modality matching methods typically focus on point-to-point distribution consistency through paired image-text sample (e.g., ranking loss based methods~\cite{zeng2020hierarchical,8692748}). Although these methods have achieved impressive results in image-text matching, they cannot effectively address human identification tasks due to the presence of many unseen identities in realistic scenarios.
However, our MFHI aims to bridge visual and semantic modalities by prototype learning and maximize the distribution consistency between the visual embedding and identity prototypes.

%------------------------------Table 1---------------------------------------------------------%
\begin{table*}[!htbp]
	%\small
	\vskip -0.18in
	\renewcommand\arraystretch{1.25}
	\centering
	\caption{Key notations.}
	\vskip -0.08in
	\begin{tabular}{c|c}
		\hline
		\textbf{Notations} & \textbf{Descriptions} \\
		\hline
		$Y^{s},Y^{u}$   & Set of seen identities and set of unseen identities, respectively \\
		\hline
		$\bm A^{tr/te},\bm {\hat{A}}^{tr/te}$ &  Set of image-level and category-level attribute descriptions about training/testing identities, respectively \\
		\hline
		$\bm X^{tr}, \bm X^{te},\bm X^{q}$ & Set of training images, set of testing images, and set of query images, respectively \\
		\hline
		$K,L$ & Number of seen identities and unseen identities, respectively \\ 
		\hline
		$\bm x_{i}^{tr},\bm y_{i}^{tr}$ & The $i$-th labeled training image: $\bm x_{i}^{tr} \in \bm X^{tr}, \bm y_{i}^{tr} \in Y^{s}$ \\
		\hline
		$\bm x_{i}^{te}$ & The $i$-th unlabeled testing image: $\bm x_{i}^{te} \in \bm X^{te}$ \\
		\hline
		$\bm x_{i}^{q}$ & The $i$-th query image: $\bm x_{i}^{q} \in \bm X^{q}$ \\
		\hline
		$\bm a_{i}^{tr/te}$, $\bm \hat{a}_{j}^{tr/te}$ & The attribute vector of the $i$-th training/testing image, and the attribute vector of the $j$-th training/testing identity\\
		\hline
		$f_{\text{I2A}}(\cdot)$ & The function of classifying an image by attribute descriptions \\
		\hline
		$f_{\text{A2I}}(\cdot)$ & The function of retrieving images with the same identity as an attribute query \\
		\hline
		$f_{\text{I2I}}(\cdot)$ & The function of predicting the correct label for an image query\\
		\hline
		
	\end{tabular}
	\vskip -0.15in
	\label{table:key_notations}
\end{table*}
%------------------------------Table 1---------------------------------------------------------%

We emphasize our \textbf{contributions} in four aspects:
\begin{itemize}
	\item[-] To the best of our knowledge, our work describes the first algorithmic framework for the modality-free human identification task.
	%  We propose for the first time modal free face identification problem, which studies more common application scenarios in real life.
	\item[-] We formulate such a task as a generic zero-shot learning model, which can bridge the visual and semantic modalities by prototype learning.
	%  We formulate modal free face identification problem as zero-shot learning problem which aims to learn transfer knowledge from seen person to target person.
	\item[-] To obtain highly discriminative features in both global category-level and local attribute-level, a semantics-guided attention map is further learned in our model, thus leading to more reliable and interpretable identification.
	%  To enhance the discriminative ability of learned features, we propose a \textit{Attribute-Driven Spatial Attention Mechanism} to capture most discriminative regions in the image.
	\item[-] Extensive experiments demonstrate our method could achieve competitive results on modality-free identification, including face identification task and re-ID task.
	%  Extensive experiments demonstrate our method can effectively address modal free face identification on two benchmarks: CelebA\cite{liu2015deep} and LFWA\cite{liu2015deep}
\end{itemize}
%------------------------------Introduction---------------------------------------------------------%
% into correct a small scale categories within a set semantic attribute descriptions.

%------------------------------Relate work---------------------------------------------------------%
\section{Related Work}

\subsection{Human Identification}
\textbf{Face Identification.}
From the view of designing the loss functions, existing face identification in open-set methods can be divided into two categories.
The first group, such as softmax based methods~\cite{cao2018vggface2}, views each identity as a category, and then trains a multi-class classifier to classify different identities.
The other group directly learns an embedding, such as~\cite{schroff2015facenet} that used the triplet loss to separate the positive pair (i.e., two images with the same identity) from the negative by a distance margin.
To further enhance the discriminability of visual features, there also appear some variants of the softmax loss recently~\cite{deng2019arcface,xu2021consistent}.
% For example, \cite{wang2018cosface} reformulated the softmax loss as a cosine loss by $\ell_{2}$ normalizing both visual features and weight matrix to eliminate radial variations.
For example, Deng et al.~\cite{deng2019arcface} proposed an additive angular margin loss to directly optimize the geodesic distance margin by virtue of the exact correspondence between the angle and arc in the normalized hypersphere.
This can help to obtain highly discriminative features for face identification.
Benefiting from the large-scale training data and the elaborate CNN, both the softmax-loss-based and the triplet-loss-based methods can achieve excellent performances on face identification.
However, all these methods only consider an image modality, which is seriously limited for the scenarios where only a textual query or an attribute gallery set is available.

%However, there still exist three main drawbacks.
%First, a significant training cost increase is caused with increasing the identities number. This mainly lies in the linear transformation matrix of softmax loss, or a combinatorial explosion in the number of face triplet.
%Second, these models all only consider an image modality, which is seriously limited for the scenarios where only a textual query or an attribute gallery set is available.
%Third, some methods, such as softmax-loss-based, can only learn the features that are separable for the close-set classification problem but not discriminative enough for the open-set.
%In this work, we aim to address all these challenges, i.e., large-scale, open-set, and modality-free face identification.

\textbf{Person Re-identification.}
Most re-ID researches take images as probes (i.e., queries) and then retrieve the corresponding person images from a pre-collected gallery set~\cite{zeng2020hierarchical}.
However, due to the limitations of realistic scenarios, we are not always able to obtain the query images.
Recently, a number of researches focus on language-based person re-ID~\cite{lin2019improving,10.1145/3383184,chai2021video,dong2019person}.
These language-based re-ID methods can be divided into two categories.
The first category~\cite{lin2019improving,chai2021video} aims to learn a more discriminative visual representation with the help of attributes
For example, a multi-task person re-id network was proposed in ~\cite{lin2019improving} to learn a Re-ID embedding and at the same time to predict the pedestrian attributes. 
Tianrui Chai \textit{et al}.~\cite{chai2021video} proposed an Attribute Salient Region Enhance (ASRE) module to learn a better separation of the pedestrian from background.
The second category~\cite{10.1145/3383184,dong2019person} aims to retrieve the matched person images when only a textual description is given, such as in~\cite{10.1145/3383184}, an instance loss for instance-level image-text matching was proposed based on the assumption that each image/text group is distinct. 

However, there are still existing additional challenges in various application scenarios.
First, existing methods are not flexible, since they can only solve queries with single text or image modality, so it is difficult to extend them to modality-free queries. 
Second, it is expensive collect rich natural language description for each person images.
Third, some methods based on natural language descriptions are discommodious because they usually need to model rich and complex sentence syntax.
%Meanwhile, the images in re-ID are usually with poor quality, which brings additional difficulties.
In contrast, collecting short attribute descriptions is very simple, and it also retains most of the semantic information.
Specially, in realistic scenarios, short attribute descriptions make it easier for witnesses to paint a portrait of the criminal, and if the search results do not meet expectations, we can quickly and accurately modify the descriptions to refine the search results.
%The goal of this work is to design a framework that is not constrained by modalities, and meanwhile reduces computational costs.

%--------------------Figure 3---------------------------%
\begin{figure*}[!htbp]
	\vskip -0.2in
	\begin{center}
		\includegraphics[width=0.6\textwidth]{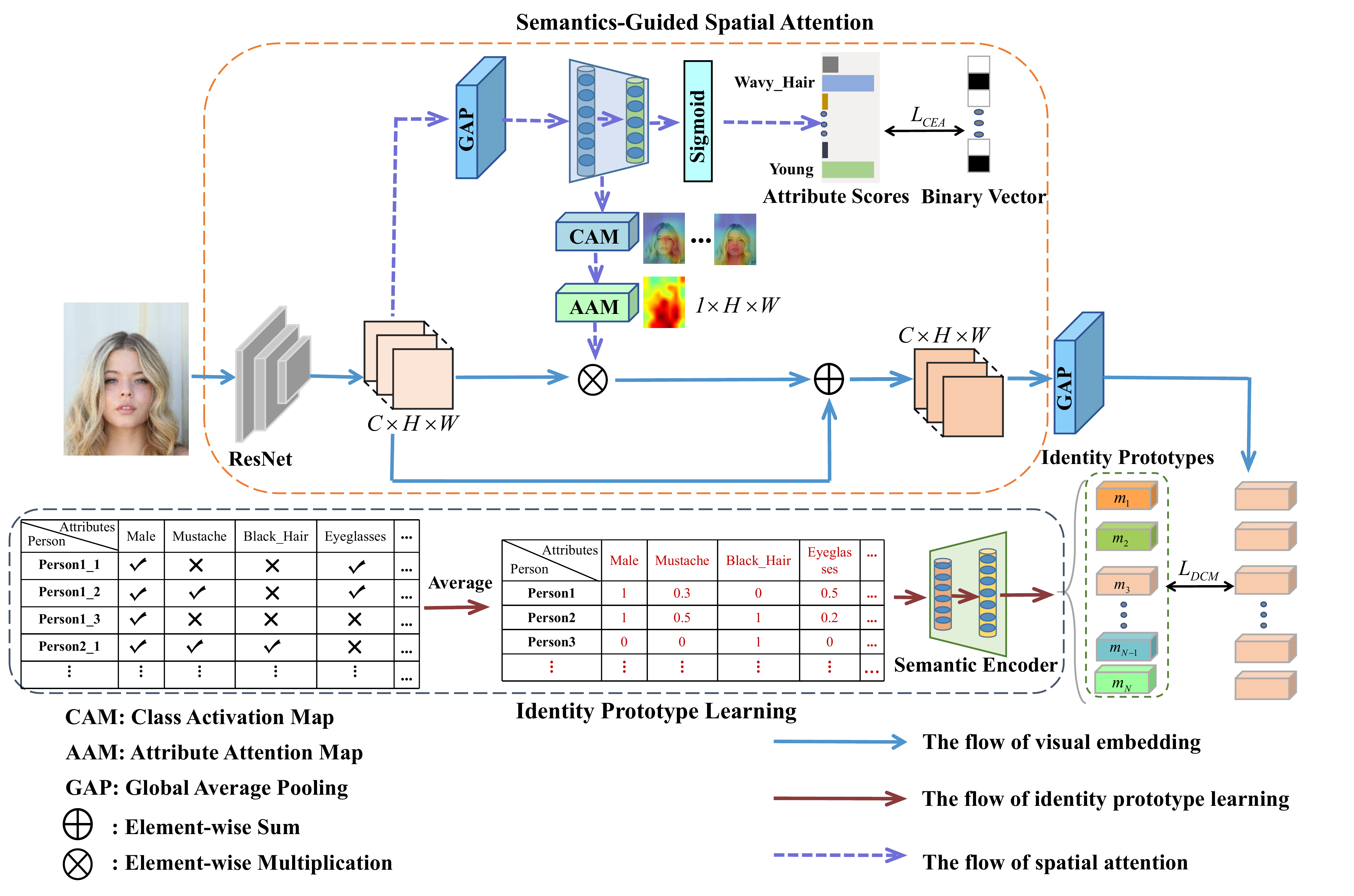}
	\end{center} %\linewidth
	\vskip -0.1in
	\caption{The overall architecture of our proposed generic zero-shot learning model for modality-free human identification.}
	\vskip -0.15in
	\label{fig:framework}
\end{figure*}
%--------------------Figure 3---------------------------%

% \textbf{Zero-Shot Learning.}
\subsection{Zero-Shot Learning}
Existing ZSL models consist of transductive~\cite{8931651} and inductive~\cite{zhang2017learning,annadani2018preserving,Xie_2019_CVPR,LIU2020103924} two settings, with the former employing the information about the unseen domain during model training, and the latter not.
Specially, the transductive setting in ZSL is proposed to alleviate the domain shift problem between seen and unseen domains.
% Specially, the transductive setting in ZSL is proposed to alleviate the domain shift problem between seen and unseen domains. It is usually embodied by two progressive degrees: transductive for specific unseen classes~\cite{liu2018generalized} and transductive for specific testing samples~\cite{8678663}.
Despite the transductive ZSL approaches have been shown to improve the generalizability of the model in the unseen domain, it is not always feasible in realistic scenarios, e.g., open-set setting in face identification.
Thus, we take an inductive setting in this work.

From the view of constructing the visual-semantic interactions, existing inductive ZSL methods can be divided into four categories.
The first group focuses on learning a projection function from the visual to the semantic space with a linear~\cite{annadani2018preserving} or a non-linear model~\cite{Xie_2019_CVPR}.
In contrast, the second group learns a projection function from the semantic to the visual space~\cite{zhang2017learning,chen2021free}.
In~\cite{chen2021free}, Chen \textit{et al}. developed a feature refinement (FR) module that incorporates semantic$\rightarrow$visual mapping into a unified generative model to refine the visual features of seen and unseen class samples.
%In~\cite{zhang2017learning}, Zhang et al. learned a deep ZSL model generating a representative visual prototype for each unseen class based on the semantic description.
%Meanwhile, it proved that during the testing phase the subsequent nearest neighbour search in the visual space would suffer much less from the hubness problem.
Combining the above methods, the third group employed an encoder-decoder paradigm with the visual or class prototype reconstruction constraint~\cite{annadani2018preserving}.
For example, by utilizing the structure of the space spanned by the attributes, PSR~\cite{annadani2018preserving} learned an encoder-decoder multilayer perceptron (MLP) model to preserve the structure of the semantic space in the visual space.
The last group learns an intermediate space, where both the visual and semantic space are projected to~\cite{chen2021hsva}.

\section{Methodology}
In this section, we first set up our modality-free human identification problem, and then formulate it as a generic zero-shot learning model.

% \textcolor{red}{ In this section, we first formulate the modal free face identification as zero-shot learning recognition problem~(Section~\ref{Setup}), then develop a \textit{Attribute-Driven Spatial Attention} based convolution prototype learning framework for this task, and finally introduce how conduct identity/retrieval on test target person.}

\subsection{MFHI: Problem Definition}\label{Setup}
The key notations used throughout this paper are summarized in Table~\ref{table:key_notations}.
Let $Y^{u}=\left \{u_1, \cdots, u_L\right \}$ denote the unseen target identity set, and these $L$ identities do not have any labeled training samples.
However, we have an attribute vector set $\bm A^{te}=\left [\bm a_{1}^{te},\cdots,\bm a_{T}^{te}\right ]$ (i.e., an attribute gallery set), where $\bm a_{i}^{te} \in \mathbb{R}^Q$ is a binary vector. It involves the textual descriptions of $T$ images about the $L$ identities.
Then we can get the category-level attribute vector set $\bm {\hat{A}}^{te}=\left [\bm {\hat{a}}_{1}^{te},\cdots,\bm {\hat{a}}_{L}^{te}\right ]$ for these $L$ identities by averaging the image-level attribute vectors of each identity.
Similarly, an image gallery set is denoted as $\bm X^{te}=\left [\bm x_{1}^{te},\cdots,\bm x_{T}^{te}\right ]$, which represents $T$ images belonging to the $L$ identities.
In addition, an image probe set consisting of $P$ queried images about the $L$ identities is represented as $\bm X^{q}=\left [\bm x_{1}^{q},\cdots,\bm x_{P}^{q}\right ]$.
% During the test phase, $\bm {\hat{A}}^{te}$ and $\bm A^{te}$ are image-level and category-level attribute vector sets of test samples respectively.
During the testing phase, we denote
(i) \textbf{Image$\rightarrow$Attribute (I2A)} as the task that assigns a correct label for a testing image of the target identity when only an attribute gallery set is available.
Therefore, we need to learn a function $f_{\text{I2A}}(x, \bm {\hat{A}}^{te})$ to predict the label $y$ of an image $x$ within $L$ different identities.
(ii) While a queried textual description $\bm a$ is provided, \textbf{Attribute$\rightarrow$Image (A2I)} focuses on retrieving the images with the same identity from an image gallery set.
Thus, a retrieval function $f_{\text{A2I}}(\bm a, \bm X^{te})$ is required.
(iii) As a common setup in conventional human identification, 
\textbf{Image$\rightarrow$Image (I2I)} aims to verify the label for a queried image $x$ from a probe set based on an image gallery set.
For this end, we need to learn a prediction function $f_{\text{I2I}}(x, \bm X^{te})$.

To flexibly and effectively tackle the above three tasks, as the previous human identification work~\cite{deng2019arcface}, we also introduce a training set $\mathcal{D}^{tr}=\left\langle\mathcal{X}^{tr},\mathcal{Y}^{tr}\right \rangle$ that contains another $K$ identities, where $\mathcal{X}^{tr}$ and $\mathcal{Y}^{tr}$ denote the sets of training images and labels respectively. 
Let $Y^s=\left \{s_1,\cdots,s_K\right \}$ denote the set of these $K$ identities. Likewise, $\bm A^{tr}=\left [\bm a_{1}^{tr},\cdots,\bm a_{S}^{tr}\right ]$ and $\bm {\hat{A}}^{tr}=\left [\bm {\hat{a}}_{1}^{tr},\cdots,\bm {\hat{a}}_{K}^{tr}\right ]$ represent their image- and category-level attribute vector sets respectively.
It is worth noticing that, typically, $K>L$, and $Y^{u} \cap Y^{s} = \emptyset$.
This is just our open-set modality-free human identification problem.

% We use a convolutional neural network denoted as  $\Psi \left( \cdot \right )$ to learn feature representation for any input image $x_i$.
% Furthermore, let $f\left ( \Psi \left( x_i \right ), \bm {\hat{A}}^{te}, \bm A^{te} \right ): \mathbb{R}^d\to\mathbb{R}^L$ denote a classifier to predict the label $y$ of the test image $x_i$ within $L$ different identities.
% $y_{i_{true}}$ represents the true label of $x_i$.
% Finally, we can represent the cost function of $f\left ( \Psi \left( x_i \right ), \bm {\hat{A}}^{te}, \bm A^{te} \right )$ as $\mathcal{L}\left( x_i,\bm {\hat{A}}^{te}, \bm A^{te},y_{true}\right)$.

\subsection{MFHI: Overall Framework}
The overall framework of our MFHI is shown in Fig.~\ref{fig:framework}, which mainly includes three flows (visual embedding flow, spatial attention flow, and prototype learning flow). Note that it is a unified architecture that can be flexibly applied to learn the three kinds of prediction functions, i.e., $f_{\text{I2A}}(\cdot)$, $f_{\text{A2I}}(\cdot)$, and $f_{\text{I2I}}(\cdot)$ in all three scenarios.
Concretely, for any input image $x$, the \emph{visual embedding flow} first obtains a visual feature map $F=\Psi \left( x \right )\in \mathbb{R}^{C\times H\times W}$.
Afterwards, the feature map $F$ is fed into the \emph{spatial attention flow} to generate an attention map.
Specially, in this flow, we first infer the attribute scores which help to gain the confidence about each attribute.
Then, for each attribute, we build the class activate map which can localize the class-sensitive activation region.
Relying on the attribute scores, we select the \textit{Top-D} attribute class activation maps of the image, since the attentive regions captured in these activation maps have important discriminability.
After that, the attribute attention map $M_{a} \in \mathbb{R}^{1\times H\times W}$ is generated by maximum operation to these activation maps.
Consequently, the new feature map $F^{'}=\left( F\otimes M_{a}\right ) \oplus F$ is updated by enforcing such a semantics-guided attention map, where $\otimes$ and $\oplus$ represent element-wise multiplication and element-wise sum operations respectively.
Thus, we can get the more discriminative visual feature of the input image via global average pooling operation to $F^{'}$.
Furthermore, a prototype learning flow is introduced to learn an identity prototype from its category-level textual description via a MLP. As a result, we can perform MFHI with modality-free property, since the visual and semantic spaces can be bridged seamlessly by the three flows.
Once the visual feature of an image and the class prototype of a target identity are learned, we can conduct modality-free human identification.

\subsection{MFHI: Semantics-Guided Spatial Attention (SGSA)}
Existing ZSL methods mainly focus on extracting global visual feature from an image.
This is usually ineffective for large-scale human identification, where there exist small inter-class and large intra-class distances.
To maximize inter-class separability, some methods~\cite{tay2019aanet,yang2021sega,ge2021semantic} proposed semantic-guided attention to capture local discriminative regions.
Specially, inspired by AAnet~\cite{tay2019aanet}, we propose to learn a semantics-guided spatial attention (SGSA) map to capture the most discriminative local attribute regions.
Meanwhile, such an attention can lead to more reliable and interpretable identification.
It is worth noting that our SGSA is greatly different from AAnet especially in network complexity and input.
Specifically, our SGSA consists of an attribute prediction module (APM) that can individually predict each attribute and an attribute attention module (AAM). Referring to~\cite{zhou2016learning}, AAM can produce a class activation map for each individual attribute by utilizing the weight of attribute classifier in APM.
Then, according to the predicted attribute scores of an image. the semantics-guided spatial attention map is generated by aggregating the \textit{Top-D} class activation maps.
% Referring to \cite{zhou2016learning}, the second sub-module utilizes the weight of attributes prediction and the feature map $F$ to generate class activation map (CAM) for each individual attribute.
% The CAM can highlight the class-specific discriminative regions.
% After that, we select the class activation maps of the most likely possessing attributes in the image to generate the final attention map.
% We introduce these two sub-modules in detail below.
\begin{itemize}
	\item[\textbf{-}] \textbf{Attribute Prediction Module (APM)}
\end{itemize}
% The key of this module is to effectively and individually predict each attribute.
% Different from normally multi-task learning based methods, we unified multi classifiers to a simple multilayer perceptron (MLP) network by designing the corresponding loss function.
% It is capable of reducing the model complexity meanwhile keeping significant performance.
As shown in Fig.~\ref{fig:framework}, the feature map with the size of $C\times H\times W$ is first down-sampled to $C \times1\times1$ by global average pooling (GAP) operation, which helps find all discriminative regions in an image.
We reshape it to $\mathbb{R}^{C\times1}$ to get the visual feature vector.
Then,  we can predict the attribute score vector $p\in \mathbb{R}^{Q}$ based on the locally discriminatively enhanced visual features.
After that, $p$ is normalized by a sigmoid activation layer, and finally we can get the confidence score of each attribute.
% There are several differences between our SGSA and AANet:
% \textit{First,} AANet chooses multi-task learning mechanism to predict attribute, which needs to build a very complex network.
% However, we unified the multi classifiers to a simple multilayer perceptron (MLP) by designing the effectively loss function and rational selection mechanism.
% \textit{Second,} the face attributes are usually more fine-grained than human attributes.
% To this end, we don't split feature maps for specially part and directly  use all channel feature maps to predict each individual attributes.
% The SGSA consists of two sub-modules (i) attribute prediction module and (ii) attribute attention modulde (AAM) generation.
\begin{itemize}
	\item[\textbf{-}] \textbf{Attribute Attention Module (AAM)}
\end{itemize}
To obtain the representation with local attribute-level discrimination, we first get the class activation map for each individual attribute.
Let $\bm W_{apm}\in \mathbb{R}^{Q\times C}$ represent the weight matrix of attribute classifier in APM, and $w_{ij}$ is the $j$-th column of $i$-th row in $\bm W_{apm}$.
Then we define $CAM_i$ as the class activation map of the $i$-th attribute, where the element in spatial position $(a,b)$ is computed as follows:
\begin{align}
\begin{split}
CAM_i\left (a,b \right ) = \sum_{j=1}^C w_{ij} F_j\left (a,b\right )
\end{split}
\end{align}
where $a\in[1,H]$ and $b\in[1,W]$. $F_j$ represents the $j$-th channel of feature map $F$.

Meanwhile, we descendingly sort the attribute scores $p$ obtained in APM.
Then, we select the \textit{Top-D} attributes with the highest scores, since these local regions normally have significant representation ability.
After that, we combine the CAMs of these attributes by maximum operation to generate the spatial attention map $M_a\in \mathbb{R}^{1\times H \times W}$.
The new feature map with both global category-level and the local attribute-level discrimination $F'\in \mathbb{R}^{C\times H \times W}$ is generated by $M_a$ as:
\begin{align}
\begin{split}
F' = (F \otimes M_a) \oplus F
\end{split}
\end{align}
Finally, we can get the final visual feature $v \in \mathbb{R}^{C\times1} $ by global average pooling operation on $F{'}$, thus leading to more reliable and interpretable identification.

% \textcolor{red}{To capture the discriminative regions, we only focus on the most likely possessing attributes in the image because these local information can effectively improve the representation ability of the learned feature.}
% To this end, we descendingly sort the attribute score $p$ and select \textit{Top-D} attribute class activation maps to generate the final spatial attention map $M_s\in \mathbb{R}^{1\times H \times W}$.

\subsection{MFHI: Prototype Learning Module (PLM)}
Generally, there often exist noises in a list of textual descriptions of one target identity. 
To address this problem, we first average them to obtain a category-level attribute vector $\hat{a}_{j}$.
Moreover, the fundamental challenge in MFHI actually lies in the heterogeneity of different modalities (e.g., images and texts).
For this end, a shared space is learned to bridge the visual and semantic spaces.
Specifically, we learn an identity prototype $m_j$ from the category-level attribute vector $\hat{a}_j$ with a MLP, which is defined as $m_{j} = \Phi \left( \hat{a}_{j} \right)$,
where the dimension of the identity prototype is equal to the visual feature.
Specially, for \textbf{I2I} task, since both probe set and gallery set are composed of images, the identity prototypes are replaced by the weight of a fully connected layer as in~\cite{deng2019arcface}, and a row of the weight represents an identity prototype.
This is mainly due to the fact that during the testing phase, no additional semantic descriptions are available in \textbf{I2I} task for generating identity prototypes, while the PLM based on semantic descriptions during the training phase will bring the semantic gap problem between visual and textual modalities, thus learning low-quality visual features. 
Finally, in order to improve the discrimination of the identity prototypes, we constrain the distribution consistency of textual prototypes and visual features by enforcing an angular margin loss as in~\cite{deng2019arcface}.

\subsection{MFHI: Loss Functions}
It has been proved that episodic training can effectively mitigate the sample distribution gap between the seen and unseen classes~\cite{LIU2020103924}, i.e., the identities in training set and gallery set.
Thus, we build a series of zero-shot tasks in training set as follows by simulating the target test task for episodic training:
\begin{align}
\begin{split}
\left \{ \left \langle \mathcal{X}_{1}^{tr},\bm {\hat{A}}_{1}^{tr},\bm A_{1}^{tr}, \mathcal{Y}_{1}^{tr} \right\rangle,\cdots,\left \langle \mathcal{X}_{n}^{tr},\bm {\hat{A}}_{n}^{tr},\bm A_{n}^{tr}, \mathcal{Y}_{n}^{tr} \right \rangle \right \}
\end{split}
\end{align}
In this setup, the objective of our MFHI is defined as:
\begin{align}\label{all_loss}
\begin{split}
\Pi = \arg \min_{\Theta}\sum_{i=1}^{n}\sum_{\{x,y\}\in\{\mathcal{X}_{i}^{tr},\mathcal{Y}_{i}^{tr}\}} \mathcal{L}\left( \Theta; x,\bm {\hat{A}}_{i}^{tr}, \bm {A}_{i}^{tr},y\right)
\end{split}
\end{align}
where $\Theta$ denotes the parameter set in visual embedding, spatial attention, and prototype learning flows. %$y$ is the label of $x$.

To effectively and accurately recognize any queried sample, we design two modules which can achieve two goals of: (i) Minimizing the \underline{c}lassification \underline{e}rror of each individual \underline{a}ttribute (CEA); (ii) Maximizing the \underline{d}istribution \underline{c}onsistency between different \underline{m}odalities via learned identity prototypes (DCM).
Based on the above two objectives, we can decompose the objective in Eq.~(\ref{all_loss}) into two functions as
\begin{align}
\begin{split}
&  \mathcal{L}\left( x,\bm {\hat{A}}_{i}^{tr}, \bm {A}_{i}^{tr},y\right) \triangleq \mathcal{L}_{\text{CEA}} \left (x,\bm{A}_i^{tr},y\right ) + \mathcal{L}_{\text{DCM}} \left (x,\bm{\hat{A}}_i^{tr},y\right ) \nonumber
\end{split}
\end{align}

\textbf{Classification Error of Attributes (CEA).}
To make the class activation map accurately localize the sensitive region for each individual attribute,
we need effectively predict each individual attribute for the input image.
% The normal methods usually select multi-task learning mechanism for this task which results in the model have very high complexity.
For this end, we can minimize the classification error in the attribute level as:
\begin{align}
\begin{split}
\mathcal{L}_{\text{CEA}} \left (x,\bm{A}_i^{tr},y\right ) = \sum_{j=1}^Q -\left( r_{j}\log p_{j} +(1-r_{j})\log \left( 1- p_{j}\right )\right) 
\end{split}
\end{align}
where $r_{j}=1$ if the image possesses $j$-th attribute and 0 otherwise. $p_{j}$ is the predicted score of $j$-th attribute.

\textbf{Distribution Consistency between Modalities (DCM).}
Suppose there exist $N$ identities $Y_i^{s}= \left \{ s_{i_{1}},\cdots,s_{i_{N}} \right \}$ in the $i$-th task $\left \langle \mathcal{X}_{i}^{tr},\bm {\hat{A}}_{i}^{tr},\bm A_{i}^{tr}, \mathcal{Y}_{i}^{tr}\right \rangle$.
Based on the category-level textual descriptions, we first obtain the prototypes of all identities in this task, represented as $M=\left \{ m_{1},\cdots,m_{N} \right \}$.
Once the identity prototype set $M$ and the visual feature $v$ of the input image are obtained, we then normalize them with $\ell_2$ operation that can make the predictions only lie on the angle between the image feature and the identity prototype vectors.
Finally, the visual features are distributed on a hypersphere with a radius of $r$.
Here, the probability of the input image $x$ belonging to the $j$-th prototype $m_j$ can be represented as:
\begin{align}
\begin{split} \label{probability}
p\left(x\in m_j \right|x) = \frac{\exp^{r\cos\theta_j}}{\sum_{l=1}^{N}\exp^{r \cos\theta_l}}
\end{split}
\end{align}
where $\theta_{j}$ represents the angle between $v$ and $m_{j}$. Moreover, to enhance the intra-class compactness and the inter-class discrepancy, there adds an angular margin $d$ penalty in the normalized hypersphere.
Eq.~(\ref{probability}) is then reformulated into:
\begin{align}
\begin{split} \label{dcm}
p\left(x\in m_j \right|x) = \frac{\exp^{r \cos\left (\theta_j+q_j\cdot d \right )}}{\exp^{ r\cos\left(\theta_{y}+d \right )}+\sum_{l=1,s_{i_{l}}\neq y}^{N}\exp^{r\cos \theta_l}}
\end{split}
\end{align}
where $\theta_{y}$ represents the angle between $v$ and the prototype with the same identity (i.e., $m_y$). $q_j=1$ if $s_{i_j} = y$ and 0 otherwise. 
Obviously, compared with the ranking loss based methods, the constraint of DCM in our MFHI can significantly enhance the inter-class separability. Unlike the previous triplet loss based methods~\cite{zeng2020hierarchical} that consider only one positive pair and one negative pair, our DCM constraint constructs one positive pair and multiple negative pairs at the same time through prototype learning, which will be more conducive to insight into the distributions of different classes, and the separability between classes can be significantly enhanced.
Finally, a promising model can be obtained by conducting a cross entropy loss as
\begin{equation}
\mathcal{L}_{\text{DCM}} \left (x,\bm{\hat{A}}_i^{tr},y\right ) = -\sum_{j=1}^{N}q_j\log p \left( x\in m_j|x \right).
\end{equation}

%------------------------------Table 2---------------------------------------------------------%
\begin{table*}[!htbp]
	%\scriptsize
	\vskip -0.2in
	\centering
	\caption{Comparative results (\%) on two datasets under two scenarios. \textcolor{red}{Red}/\textcolor{blue}{Blue} represent the Best/second best results.}
		\vskip -0.05in
	\begin{tabular}{|c|c|c|c|c|c|c|c|c|c|c|c|c|c|c|c|c|}
		\hline
		\multirow{3}*{Method Categories} & Datasets & \multicolumn{6}{|c|}{CelebA} & \multicolumn{6}{|c|}{LFWA} \\ \cline{2-14}
		&	\multirow{2}*{Methods} 	& \multicolumn{3}{|c|}{\textbf{Image$\rightarrow$Attribute}} & \multicolumn{3}{|c|}{\textbf{Attribute$\rightarrow$Image}} & \multicolumn{3}{|c|}{\textbf{Image$\rightarrow$Attribute}} & \multicolumn{3}{|c|}{\textbf{Attribute$\rightarrow$Image}} \\ \cline{3-14}
		&	& Top-1 & Top-5 & Top-10 & R@1 & R@5 & R@10 & Top-1 & Top-5 & Top-10 & R@1 & R@5 & R@10  \\
		\hline
		\multirow{3}*{ZSL} & DEM~\cite{zhang2017learning}  & 6.0  & 18.5 & 27.9 &  \textcolor{blue}{21.4} & 41.5 & 54.6 & 4.0 & 11.7 & 17.5 & 25.7 & 48.7 & 60.1    \\
		&	RN~\cite{sung2018learning}    & 0.5 & 2.5 & 5.0 & 0.1 & 2.2 & 4.1 & 0.5 & 1.1 & 1.8 & 0.2 & 0.6 & 1.1   \\
		&	AREN~\cite{Xie_2019_CVPR}     & \textcolor{blue}{24.3} & \textcolor{blue}{51.5} & \textcolor{blue}{63.8} & 3.6 & 10.6 & 14.9 & \textcolor{blue}{21.9} &  \textcolor{blue}{47.2} &  \textcolor{blue}{59.9} & 1.3 & 4.1 & 6.6   \\
		\hline
		\multirow{2}*{GCL} & VSE++~\cite{faghri2017vse++}   & 6.6 & 20.0 & 29.3 & 10.8 & 28.5 & 39.8 & 5.3 & 17.2 & 26.7 & 4.5 & 13.4 & 20.2   \\
		&	VSRN~\cite{li2019visual}    & 9.5 & 25.5 & 35.2 & 16.9 & 38.0 & 49.9  & 7.4 & 21.0 & 30.5 & 6.4 & 18.3 & 27.2   \\
		\hline
		\multirow{2}*{LAL} & SCAN~\cite{lee2018stacked}    & 4.7 & 14.6 & 21.8 & 6.4 & 18.9 & 28.5 & 4.6 & 11.7 & 16.6 & 2.6 & 9.0 & 14.0 \\
		& AttPre  & 15.2 & 34.4 & 46.3 & 20.5 &  \textcolor{blue}{45.9} &  \textcolor{blue}{57.0} & 21.7 & 41.5 & 53.1 & \textcolor{blue}{26.3} & \textcolor{blue}{51.6} & \textcolor{blue}{62.5} \\
		\hline
		GLL & MFHI  & \textcolor{red}{30.1} & \textcolor{red}{58.3} & \textcolor{red}{71.0} & \textcolor{red}{41.0} & \textcolor{red}{67.7} & \textcolor{red}{77.3} & \textcolor{red}{36.3} & \textcolor{red}{65.9} & \textcolor{red}{77.1} & \textcolor{red}{39.1} & \textcolor{red}{66.7} & \textcolor{red}{76.2}   \\
		\hline
	\end{tabular}
	\vskip -0.2in
	\label{table:comparative_results}
\end{table*}
%------------------------------Table 2---------------------------------------------------------%

\subsection{MFHI: Recognition}
For {\textbf{I2A}} and {\textbf{A2I}}, there exists a given testing set $\{ X^{te}, A^{te},\bm {\hat{A}}^{te} \}$ with the unseen class set $Y^u$. 
Specially, for {\textbf{I2A}} scenario, to correctly classify any image sample in $X^{te}$, 
based on the category-level textual descriptions, the prototypes of all target identities covered in $Y^{u}$ are first generated by the learned PLM.
Then, the testing sample $x^{te}$ is embedded into the shared space to obtain the visual feature $v$.
After that, the similarity scores between the visual feature $v$ and every identity prototype are computed by cosine metric.
Finally, the testing sample is classified to the nearest prototype.
Let $y^{te}$ denote the predicted label, which can be defined as:
\begin{align}
\begin{split} 
y^{te} = \arg \max \limits_{u_{k} \in Y^{u} } p \left(v \in m^{te}_{k}|x^{te}\right)
\end{split}
\end{align}
where $m^{te}_{k} = \Phi \left( \hat{a}^{te}_{k}\right) $ is the $k$-th target identity prototype and $p \left(v \in m^{te}_{k}|x^{te}\right) = \frac{v^{T}\cdot m^{te}_{k}}{\left\|v\right\|_2\left \|m^{te}_{k}\right\|_2}$. 
Likewise, for {\textbf{A2I}}, the category-level attribute query is first fed into the PLM to produce the identity prototype.
Afterwards, the visual feature of each image in the gallery set is obtained by the image embedding module. 
In the end, the similarity score between the identity prototype and each image feature
is calculated by cosine metric.
The gallery images with top rank scores are considered as the possible matches.

Different from the above two scenarios, there exist an image probe set $X^q$ and an image gallery set $X^{te}$ on {\textbf{I2I}}.
Given an image query, we first obtain the visual features of the queried image and every gallery image, and meanwhile normalize them by $\ell_2$ operation. 
Then, the pairwise similarity scores between the queried image feature and every gallery image feature are computed by Euclidean distance metric. Consequently, we treat the gallery images with top rank scores as the possible matches.     
%\textcolor{red}{In order to retrieve the most similar sample to an image query, we first need to extract the visual features of all image samples in gallery set. 
%Meanwhile, we also extract the visual feature of an image query which comes from probe set.
%After that, we compute the cosine score between the image query and all image sample in gallery set.}
 
%------------------------------Method---------------------------------------------------------%

%------------------------------Experiment---------------------------------------------------------%
\section{Experimental Results and Analysis}
In this section, we evaluate the proposed method on two challenging tasks, i.e., face identification and re-ID.
%\textcolor{red}{Moreover, in order to evaluate the generalization ability of MFHI, We extend it to person %re-identification task with a large-scale unseen identity in testing, and conduct comprehensive experiments on a representative re-ID dataset, \textit{i.e.} Market1501~\cite{zheng2015scalable}.}

\subsection{Modality-Free Face Identification}

\textbf{Datasets.}
Among the most widely used datasets for face identification, we select two attribute datasets.
(i) \textbf{CelebA} is a large scale face dataset that consists of 202,599 images about approximately 10k identities~\cite{liu2015deep}.
Each image is annotated with 40 binary attributes.
(ii) \textbf{LFWA}~\cite{liu2015deep} is created based on unconstrained face dataset LFW~\cite{LFWTech}.
It contains 13,233 images of 5,749 identities, and each image is also labeled with 40 binary attributes.

\textbf{Protocols and Evaluation Metrics.}
For CelebA dataset, we adopt the standard split in \cite{liu2015deep}, with 162,770, 19,867, and 19,962 images for training, validation, and testing.
% i.e., the training data contains 162,770 images, the validation data consists of 19,867 images, and remaining 19,962 images are used as test data.
% The image number ratio of probe and gallery set is 8:2.
For LFWA dataset, we divide it by randomly choosing identities, and construct training and testing sets at a ratio of 8:2.
% The identities number ratio of train and test set is set to 8:2.
% We randomly select the facial images of eighty percent of the identity as training data, remaining facial images are used as test data.
Specially, The training and testing identities are completely disjoint in two datasets.
% In particular, probe and gallery sets are not built for LFWA since most identities just possess one facial image.
During the testing phase of \textbf{I2A} scenario, we treat a correct matching between the queried image and the gallery attribute description as a correct prediction.
It is expected that our MFHI should have high performance on both densely and sparsely populated identities.
Therefore, referring to~\cite{LIU2020103924}, we select mean average per-class accuracy at Top-P as the metric where $P \in \{1,5,10\}$.
For \textbf{A2I}, we treat the gallery images respecting a given attribute query as true matches. 
To evaluate the performance of MFHI, referring to~\cite{dong2019person}, we use the Cumulative Match Characteristic (CMC) at R@P as the metric and we set $\text{P} \in \{1,5,10\}$.
% This metric is typically used in zero-shot recognition.
% Let $N_{i_{true}}$ denote the number of correct predictions in \textit{i}-th identity, and $N_i$ represent the total number of images in \textit{i}-th identity
%It can be defined as follows:
%\begin{equation}
%\text{\ Top-P}=\frac{1}{L}\sum_{i=1}^{L}\frac{\text{\# correct predictions in}~u_{i}}{\text{\# samples in}~u_{i}}
%\end{equation}
% where $L$ represents the number of testing identities, and $P \in \{1,5,10\}$ in our experiments.
%For \textbf{Attribute$\rightarrow$Image}, we adopt the average Rank P retrieval accuracy metrics which is generally used in person search.
%It can be defined as follows:
%\begin{align*}
%Rank P=\frac{\text{\# correctly retrieved identities}}{\text{\# total identities}}.
%\end{align*}

% For \textbf{A2I}, we treat the gallery images respecting a given attribute query as true matches. 
% To evaluate the performance of MFHI, referring to~\cite{dong2019person}, we use the Cumulative Match Characteristic (CMC) at R@P as the metric and we set $\text{P} \in \{1,5,10\}$.
% It is defined as follows:
% \begin{equation}
% \text{\ R@P}=\frac{1}{L}\sum_{i=1}^{L}Acc_i
% \end{equation}
% where $Acc_i = 1$ if the Rank-P retrieval results contain the images with the same identity as the attribute query and 0 otherwise. In our experiments, we set $\text{P} \in \{1,5,10\}$.
%------------------------------Table 3---------------------------------------------------------%

\textbf{Implementation Details.}
For data preprossessing, we resize the face image to $112\times 112$, and augment the training data with random flipping.
The feature scale $r$ is selected from $\left \{ 32,64\right \}$, and the angular margin $d$ is selected from $\left \{ 0.1, 0.2, 0.3\right \}$.
For the visual embedding flow, we employ the widely used CNN in face recognition, IR-SE~\cite{hu2018squeeze}, and initialize it with the pretrained weight\footnote{\url{https://github.com/TreB1eN/InsightFace_Pytorch}}.
The output of the last convolutional layer with 512 channels is adopted as the visual feature map.
Moreover, for the MLP of the identity prototype learning flow, the size of hidden layer is set to 256, and the output size is set to 512 that is the same as the visual feature dimension.
In the spatial attention flow, the number $D$ of attribute activation maps is set to 10, and the input size of attribute classifier is 512, while the output size is equal to the number of attributes.
We select Adam optimizer where the learning rate is initialized to 5e-5 with the weight decay of 5e-4.

\textbf{Compared Methods.}
To evaluate the superiority of our joint global- and local-level (i.e., GLL) embedding approach, we compare with a wide range of plausible solutions to modality-free face identification problem.
% These methods can be divided into two categories:
(i) \textit{Global category-level visual-textual embedding} methods (i.e., GCL): Including two ZSL methods (i.e., DEM~\cite{zhang2017learning} and RN~\cite{sung2018learning}) and two visual semantic embedding methods (i.e., VSE++~\cite{faghri2017vse++} and VSRN~\cite{li2019visual}).
(ii) \textit{Local attribute-level visual-textual embedding} methods (i.e., LAL): Such as a region proposal based dense image-text cross-modal matching method SCAN~\cite{lee2018stacked} and an attentive region embedding ZSL method AREN~\cite{Xie_2019_CVPR}.  
A method based on attribute prediction (named AttPre) is designed for face identification to evaluate the identification performance of relying on individual attribute recognition.
% In addition, to evaluate the identification performance of relying on individual attribute recognition, a method based on attribute prediction (named AttPre) is designed for face identification.
For testing language models, including VSE++~\cite{faghri2017vse++}, SCAN~\cite{lee2018stacked}, and VSRN~\cite{li2019visual}, we use random attribute sentences due to lack of order, and then report the average results for 10 trails.
For all methods, we use IR-SE to extract visual features except SCAN~\cite{lee2018stacked} and VSRN~\cite{li2019visual}, while SCAN~\cite{lee2018stacked} and VSRN~\cite{li2019visual} employ the pretrained bottom up attention model~\cite{8578734} as the visual embedding module.
Moreover, it is worth noting that existing face identification methods cannot be extended to modality-free task, since they just consider a image modality.

\textbf{Experimental Results and Analysis.}
We report the comparative results in Table~\ref{table:comparative_results}.
It can be seen that:
(i) Our MFHI outperforms all existing methods on two benchmarks in two scenarios, validating that our method is very flexible and can effectively address the modality-free face identification problem.
For instance, for Top-1 of \textbf{I2A} and R@1 of \textbf{A2I}, the improvements obtained by our MFHI over the strongest competitors on two datasets range from 5.78\% to 19.60\% and 12.84\% to 14.43\%, respectively.
Such competitive results prove that the different face modalities (i.e., images and texts) possess the consistent distributions in the shared space by learning a semantic visual prototype for each identity.
% Thus, the queried samples from the probe set can be accurately identified.
(ii) The state-of-the-art ZSL methods fail to excel due to higher inter-class similarity, larger intra-class variation, larger search spaces, and more challenging application scenarios.
For example, the best ZSL method AREN~\cite{li2019visual} achieves satisfied results in \textbf{I2A}, but it significantly drops in \textbf{A2I}.
% Particularly, RN~\cite{sung2018learning} cannot generalize well to this task since it is hard to learn a suitable metric function for the image-text similarity metric in a larger search space.
(iii) Comparing with GCL methods and the LAL embedding method, we can find that there are larger improvements obtained by our MFHI. 
This indicates that these image-text matching methods which typically focus on point-to-point distribution consistency through paired image-text sample cannot effectively address the identification of unseen identities 
% require complex natural language descriptions associated with images, and these methods cannot extract useful semantic information from simple short attribute descriptions. 

%(iv) Compared with strong competitor AttPre, the improvements obtained by our method range from 14.87\% to 24.65\% on CelebA, and 11.18\% to 23.42\% on LFWA.
%This can be attributed to the global category-level discrimination with an additive angular margin, and meanwhile, the captured discriminative attribute regions via semantics-guided spatial attention.

\subsection{Modality-Free Person Re-Identification}

\textbf{Datasets.} 
For re-ID, We select two representative person search datasets: Market-1501 dataset~\cite{zheng2015scalable} and DukeMTMC-reID dataset~\cite{ristani2016performance,lin2019improving}.
%we compare our MFHI with several state-of-the-art person search methods on Market-1501 dataset~\cite{zheng2015scalable} and DukeMTMC-reID dataset~\cite{ristani2016performance,lin2019improving}.
Market-1501 contains 751 identities for training and 750 identities for testing.
The training set, gallery set, and probe set contain 12,936 images, 19,732 images, and 3,368 queried images respectively.
Meanwhile, each image is annotated by 27 attributes.
Both training and testing include 702 identities on DukeMTMC-reID.
The training set, gallery set, and probe set contain 16,522 images, 17,661 images, and 2,228 queried images respectively.
Meanwhile, each image is annotated by 23 attributes.
In order to facilitate model training, we change the original attribute vector to the one-hot vector, and then the final dimension of the attribute vector is 35 for Market-1501 and 26 for DukeMTMC-reID.
Specially, for \textbf{A2I} and \textbf{I2A} two scenarios, we re-assign semantic IDs for each person image according to its attribute vector rather than real identity, which means different people with the same attribute vector have the same semantic ID.
Finally, for Market-1501, we have 508 semantic IDs for training and 484 semantic IDs for testing, and for DukeMTMC-reID, we have 300 semantic IDs for training and 387 semantic IDs for testing.

\begin{table*}[!htbp]
	\centering
	\scriptsize
	\vskip -0.2in
	\caption{Comparative results (\%) on Market-1501 dataset and DukeMTMC-reID dataset. - represents that these methods did no release official codes, and \textcolor{red}{Red}/\textcolor{blue}{Blue} represent the Best/second best results.}
		\vskip -0.05in
	\begin{tabular}{|c|c|c|c|c|c|c|c|c|c|c|c|c|c|c|c|c|c|}
			\hline
		\multirow{3}*{Method Categories} & Datasets & \multicolumn{7}{|c|}{Market-1501} & \multicolumn{7}{|c|}{DukeMTMC-reID} \\ \cline{2-16}
		&	\multirow{2}*{Methods} 	& \multicolumn{3}{|c|}{\textbf{Image$\rightarrow$Attribute}} & \multicolumn{4}{|c|}{\textbf{Attribute$\rightarrow$Image}} & \multicolumn{3}{|c|}{\textbf{Image$\rightarrow$Attribute}} & \multicolumn{4}{|c|}{\textbf{Attribute$\rightarrow$Image}} \\ \cline{3-16}
		&	& Top-1 & Top-5 & Top-10 & R@1 & R@5 & R@10 & mAP & Top-1 & Top-5 & Top-10 & R@1 & R@5 & R@10 & mAP  \\
		\hline
		\multirow{3}*{ZSL} &	DEM~\cite{zhang2017learning}  & 22.6   &  48.9  & 59.1 & 34.0   & 48.1   & 57.5 & 17.0 & 10.7 & 28.2 & 38.6 & 22.7 & 43.9 & 54.5 & 12.9\\
		&	RN~\cite{sung2018learning}    & 22.8 & 53.0 & 65.8 & 17.2   & 38.7   & 47.3    & 15.5 & \textcolor{blue}{22.4} & \textcolor{blue}{49.2} & \textcolor{blue}{61.7}  & 25.1 & 42.0 & 51.5 & 13.0 \\ 
		&	AREN~\cite{Xie_2019_CVPR}     &\textcolor{blue}{28.8} & \textcolor{blue}{62.0} & \textcolor{blue}{76.3} & 21.9 & 38.6 & 46.1 & 15.3  & 17.5 & 39.3 & 50.8 & 23.7 & 42.2 & 50.9 & 12.8 \\
		\hline
		\multirow{8}*{GCL} &	DeepCCA~\cite{andrew2013deep}  & - &- & - & 29.9   & 50.7   & 58.1    & 17.5 & -&- & - & 36.7 & 58.8 & 65.1 & 13.5 \\ 
		&	2WayNet~\cite{eisenschtat2017linking} & -&- & - & 11.2   & 24.3   & 31.4    & 7.7  & -&- & -  & 25.2 & 39.8 & 45.9 & 10.1 \\ 
		&	MMD~\cite{tolstikhin2016minimax}   & - & - & - & 34.1   & 47.9   & 57.2    & 18.9  & -&- & - & 41.7 & 62.3 & 68.4 & 14.2\\ 
		&	DeepCoral~\cite{sun2016deep}     & - & - & - & 46.1 & 61.0   & 68.1 & 17.1  & -&- & - &  46.1 & 61.0 & 68.1 & 17.1 \\ 
		&	VSE++~\cite{faghri2017vse++}     & 15.3 & 38.7 & 51.3      & 27.0   & 49.1   & 58.2    & 17.2  & 7.9 & 21.8 & 31.1 &  33.6 & 54.7 & 62.8 & 15.5\\ 
		&	AAIPR~\cite{ijcai2018-153}       &- &- & - & 40.2   & 49.2 & 58.6 & 20.6 & -&- & -  &  46.6 & 59.6 & 69.0 & 15.6\\
		&	VSRN~\cite{li2019visual}     & 8.2 & 27.8 & 41.6  & 11.6 & 30.3 & 41.5 & 7.2  & 10.4 & 28.1 & 38.5 &  25.2 & 57.4 & 69.4 & 14.0\\ 
		\hline 
		\multirow{3}*{LAL} & SCAN~\cite{lee2018stacked} & 8.3  & 26.3 & 38.4 & 4.0    & 10.1   & 15.3    & 2.1  & 6.0 & 21.0 & 32.2 &  3.5 & 9.3 & 14.3 & 1.6\\
		&	GAN-RNN~\cite{li2017person}      & - & - & - & 30.4   & 38.7   & 44.4    & 15.4  & -&- & - & 34.6 & 52.7 & 65.8 & 14.2 \\ 
		&	CMCE~\cite{li2017identity}       & - & - & - & 35.0   & 50.9   & 56.4    & 22.8  & -&- & - & 39.7 & 56.3 & 62.7 & 15.4\\ 
		\hline
		\multirow{2}*{GLL} &	AIHM~\cite{dong2019person}       & - & - & - & \textcolor{blue}{43.3} & \textcolor{blue}{56.7} & \textcolor{blue}{64.5} & \textcolor{blue}{24.3}  & -&- & - & \textcolor{blue}{50.5} & \textcolor{blue}{65.2} & \textcolor{blue}{75.3} & \textcolor{blue}{17.4}\\ 
		&	MFHI        & \textcolor{red}{36.2} & \textcolor{red}{68.2} & \textcolor{red}{80.0} &\textcolor{red}{44.8} & \textcolor{red}{66.9} & \textcolor{red}{74.4} & \textcolor{red}{33.5} & \textcolor{red}{27.5} & \textcolor{red}{59.2} & \textcolor{red}{71.6} & \textcolor{red}{57.9} & \textcolor{red}{76.1} & \textcolor{red}{81.9} & \textcolor{red}{31.8}\\
		\hline
	\end{tabular}
	\vskip -0.2in
	\label{table:comparative_re-id}
\end{table*}
%------------------------------Table 3---------------------------------------------------------%

\textbf{Evaluation Metrics.}
As the common setups~\cite{dong2019person}, 
the gallery images respecting a given attribute/image query are considered as true matches in \textbf{A2I} and \textbf{I2I}.
We also use the CMC at R@P and the mean Average Precision (mAP)~\cite{zheng2015scalable} as our evaluation metrics, where $\text{P} \in \{1,5,10\}$ in \textbf{A2I} and $\text{P} \in \{1,10\}$ in \textbf{I2I}.
For \textbf{I2A}, we still use the mean average per-class accuracy at Top-P as the evaluation metric, where $P\in \left \{ 1,5,10\right \}$

\textbf{Compared Methods.}
For \textbf{A2I} and \textbf{I2A}, we compare our MFHI with several state-of-the-art person search methods by text attribute.
These methods can be divided into three paradigms: \textit{Global category-level} (i.e., GCL): Including CCA~\cite{andrew2013deep,eisenschtat2017linking,sun2016deep} and MMD~\cite{tolstikhin2016minimax} based cross-modal matching models, ZSL methods (i.e., DEM~\cite{zhang2017learning} and RN~\cite{sung2018learning}), visual semantic embedding methods (i.e., VSE++~\cite{faghri2017vse++} and VSRN~\cite{li2019visual}) and GAN based visual-semantic alignment method (i.e., AAIPR~\cite{ijcai2018-153}).
\textit{Local attribute-level} (i.e., LAL): Such as region proposal based matching method (i.e., SCAN~\cite{lee2018stacked}) and natural language query based person search (i.e., GAN-RNN~\cite{li2017person} and CMCE~\cite{li2017identity}).
\textit{the combination of global- and local-level visual-textual embedding methods} (i.e., GLL): Including a visual-textual hierarchical embedding GLL method (i.e., AIHM~\cite{dong2019person}). 
Additionally, in order to evaluate the advantage of our MFHI on \textbf{I2I}, we compare with a wide range of competitive and representative image query methods.

%------------------------------Table 4---------------------------------------------------------%
\begin{table}[!tbp]
	\centering
	\scriptsize
	\caption{Comparative results (\%) on Market-1501 and DukeMTMC-reID datasets. - represents that these methods did not report the corresponding results, RR denotes Re-Ranking~\cite{Zhong2017RerankingPR}, and \textcolor{red}{Red}/\textcolor{blue}{Blue} represent the Best/second best results.}
	\vskip -0.05in
	\resizebox{0.45\textwidth}{20mm}{
	\begin{tabular}{|c|c|c|c|c|c|c|}
		\hline
		\multirow{2}*{Methods} & \multicolumn{3}{|c|}{\textbf{Market-1501}} & \multicolumn{3}{|c|}{\textbf{DukeMTMC-reID}} \\ 
		\cline{2-7}
		& R@1 & R@10 & mAP & R@1 & R@10 & mAP \\
		\hline
		PDC~\cite{Su_2017_ICCV}  & 84.4 & 94.90 & 63.4 & - & - & -  \\
		DPFL~\cite{8265515}     & 88.9 & - & 73.1  & - & - & -\\
		GLAD~\cite{8466034}   & 89.9 & - & 73.9  &- & - & -\\ 
		KPM(Resnet-50)~\cite{8578818}  & 90.1 & 97.9 & 75.3  & 80.3 & - & 63.2\\ 
		PCB(Resnet-50)~\cite{Sun2018BeyondPM} & 93.8 & 98.5 & 81.6 & 83.3 & - & 69.2\\ 
		AANet(Resnet-50)~\cite{tay2019aanet}   & \textcolor{blue}{93.9} & \textcolor{blue}{98.6} & \textcolor{blue}{82.5} & \textcolor{blue}{86.4} & - & \textcolor{blue}{72.6}\\ 
		MFHI(Resnet-50)~\cite{tolstikhin2016minimax} & \textcolor{red}{95.1}  & \textcolor{red}{98.9} & \textcolor{red}{88.6} & \textcolor{red}{90.9} & \textcolor{red}{96.8} & \textcolor{red}{80.6}\\ 
		\hline
		GP-reID(Resnet-101)+\textbf{RR}~\cite{Almazn2018ReIDDR}   & 93.0 & - & 90.0 & 89.4 & - & 85.6\\ 
		SPReID(Resnet-152)+\textbf{RR}~\cite{Kalayeh2018HumanSP}   & 94.6 & 97.7 & 91.0 & 89.0 & - &85.0 \\ 
		AANet(Resnet-152)+\textbf{RR}~\cite{tay2019aanet}   & \textcolor{blue}{95.1} & \textcolor{blue}{97.9} & \textcolor{blue}{92.4} & \textcolor{blue}{90.4} & - & \textcolor{blue}{86.9}\\ 
		MFHI(Resnet-50)+\textbf{RR}  & \textcolor{red}{95.7} & \textcolor{red}{98.5} & \textcolor{red}{94.6} & \textcolor{red}{92.8} & \textcolor{red}{96.8} & \textcolor{red}{90.5}\\
		\hline
	\end{tabular}}
	\vskip -0.22in
	\label{table:comparative_re-id_i2i}
\end{table}
%------------------------------Table 4---------------------------------------------------------%

%--------------------Figure 4---------------------------%
\begin{figure*}[!tbp]
		\vskip -0.15in
	\begin{center}
		\includegraphics[width=0.7\textwidth ]{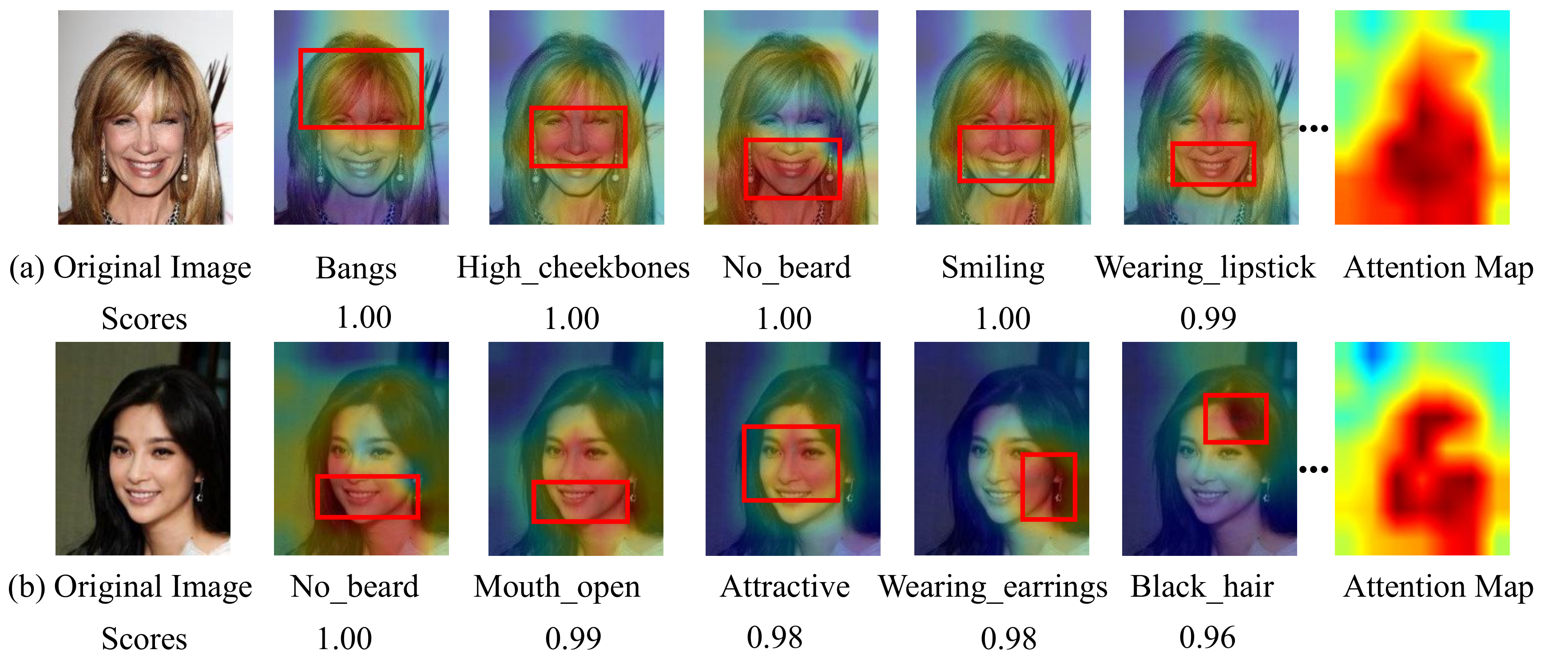}
	\end{center} %\linewidth
	\vskip -0.15in
	\caption{The visualization of class activation maps and the attribute attention map in face identification. We highlight the captured attentive region with a \textcolor{red}{red} rectangular frame in an activation map.}
	\vskip -0.1in
	\label{fig:aam_fi}
	\end{figure*}
%--------------------Figure 4---------------------------%
%--------------------Figure 5---------------------------%
\begin{figure*}[!tbp]
	\begin{center}
		\includegraphics[width=0.7\textwidth ]{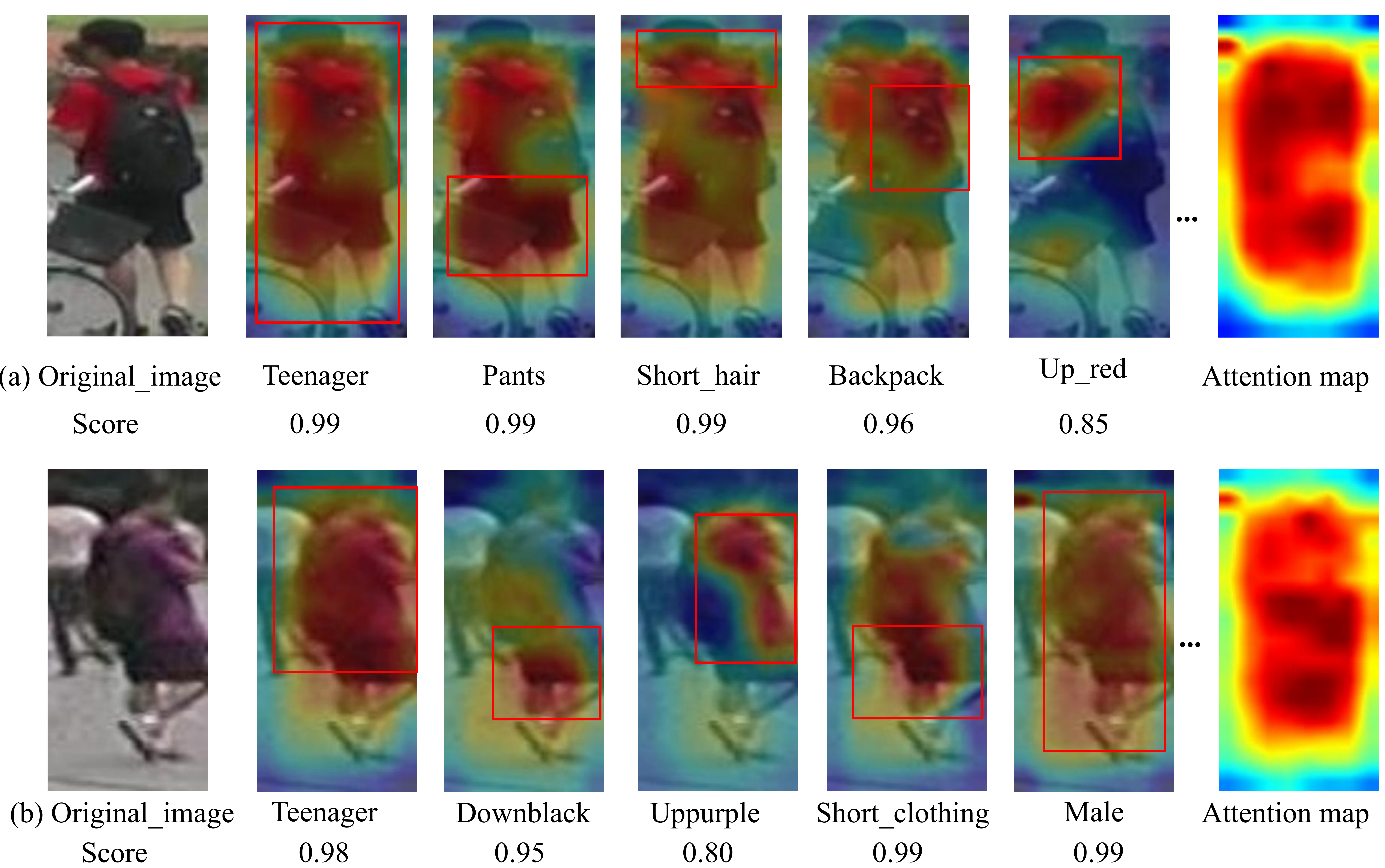}
	\end{center} %\linewidth
	\vskip -0.15in
	\caption{The visualization of class activation maps and the attribute attention map in re-ID. We highlight the captured attentive region with a \textcolor{red}{red} rectangular frame in an activation map.}
	\vskip -0.2in
	\label{fig:aam_re_id}
\end{figure*}

\textbf{Implementation Details.}
We resize the person image to $256\times 128$.
For data augmentation, we employ random flipping in \textbf{A2I} and \textbf{I2A}, and use random flipping and random erasing~\cite{zhong2017random} in \textbf{I2I}.
The  feature scale $r$ and the angular margin $d$ are selected from $\left \{8,64,128,256\right \}$ and $\left \{0.15,0.2,0.4\right \}$ respectively. 
For the visual embedding flow, we use the ResNet-50 model with instance normalization and batch normalization, which contains 2048 channels in the last convolutional layer. We initialize it with the pretrained weight\footnote{\url{https://github.com/Qidian213/Ranked_Person_ReID}}. 
Moreover, in the identity prototype learning flow, the hidden layer size of the MLP is set to 1024, and the input and output sizes are set to 35 and 2048 respectively.
In the spatial attention flow, the number $D$ of attribute activation maps is set to 8, the attribute classifier input size is 2048, and the output size is 35.
Especially, we select Adam optimizer where the learning rate is initialized to 5e-5 with the weigh decay to 5e-4 in \textbf{I2A} and \textbf{A2I}, and the initial learning rate is set to 5e-1 in \textbf{I2I}.

\textbf{Experimental Results and Analysis.}
The experimental results of re-ID are shown in Table~\ref{table:comparative_re-id} and Table~\ref{table:comparative_re-id_i2i}, respectively. About the results, we have the following discussions.

From Table~\ref{table:comparative_re-id}, it can be observed that: (i) Our MFHI model outperforms a wide variety of state-of-the-art methods, e.g., over the second best method AREN~\cite{Xie_2019_CVPR} by a margin of 7.4\% in Top-1 of \textbf{I2A} scenario on Market-1501, and over the strongest competitor person search model AIHM~\cite{dong2019person} by a margin of 7.4\% (resp. 14.4\%) in R@1 (resp. mAP) of \textbf{A2I} scenario on DukeMTMC-reID.
% These comparative results verify that our method can simply yet effectively cover such challenging re-ID tasks.
%jointly improving global- and local-level discrimination.
(ii) By comparing with the hierarchical visual-textual embedding method AIHM~\cite{dong2019person}, it can be found that capturing attribute-level discriminative regions is indeed essential for such a large-scale identification problem.
(iii) The performance margins over the GCL methods and the LAL methods are more significant.
Obviously, for such challenging large-scale tasks, modelling only from global perspective or local perspective is not the optimal solution. 
(iv) In addition, the state-of-the-art ZSL methods are hard to be extended to this problem because of its more categories, more complex background, and meaningless category names.
In contrast, our MFHI can maximize the human identity separability by minimizing the distribution discrepancy between the modalities.
% Thus, our MFHI can exactly predict the label of each query.

As shown in Table~\ref{table:comparative_re-id_i2i} for \textbf{I2I}, we can find that: (i) MFHI model yields better performances than the state-of-the-art baselines whether re-ranking~\cite{Zhong2017RerankingPR} is used or not. 
This validates that by performing a local spatial attention mechanism on visual data and maximizing the distribution consistency of identity prototypes and visual features, 
our MFHI effectively improves the discriminability of visual features.
(ii) Comparing with AANet~\cite{tay2019aanet}, which also use Resnet-50 as the visual embedding module, 
the improvements obtained by our MFHI over the best existing method are 1.2\% (resp. 6.2\%) in R@1 (resp. mAP) on Market-1501 dataset.
(iii) We also compare MFHI with AANet (Res-152)~\cite{tay2019aanet} with re-ranking (i.e., \textbf{+RR}). Our MFHI achieves 0.6\% and 2.22\% significant improvements in R@1 and mAP on Market-1501 dataset, respectively, and meanwhile has lower complexity.

\subsection{Qualitative Results Analysis}
To provide more in-depth and visual evaluations for our MFHI, we design and conduct two qualitative discussions.
First, as show in Fig.~\ref{fig:aam_fi} and Fig.~\ref{fig:aam_re_id}, we report the heatmaps of some discriminative attributes in face identification and re-ID, respectively.
This can help us to analyze the effectiveness of semantics-guided spatial attention in MFHI.
It can be found that:
(i) Our method can accurately predict each individual attribute for four images of face identification and re-ID, so as to select the most significant local regions.
(ii) In addition, Fig.~\ref{fig:aam_fi} shows that the face attributes can be exactly localized in a face image by our method, like ``bangs" in Fig.~\ref{fig:aam_fi}(a) and ``black\_hair" in Fig.~\ref{fig:aam_fi}(b).
Especially, for some local attributes that are difficult for people to observe, our model still can accurately capture the visual regions of these attributes, like ``wearing\_lipstick" in Fig.~\ref{fig:aam_fi}(a) and ``wearing\_earrings" in Fig.~\ref{fig:aam_fi}(b). 
(iii) As observed from Fig.~\ref{fig:aam_re_id}, the wear appearance of the two people is successfully captured, like ``backpack" in Fig.~\ref{fig:aam_re_id}(a) and ``uppurple" in Fig.~\ref{fig:aam_re_id}(b). Then, by extracting such representative visual regions, the local-level discriminability of the visual features can be effectively enhanced. Finally, we can obtain more reliable and interpretable identification.

\begin{figure*}[!htbp]
	\vskip -0.1in
	\begin{center}
		\includegraphics[width=0.75\textwidth ]{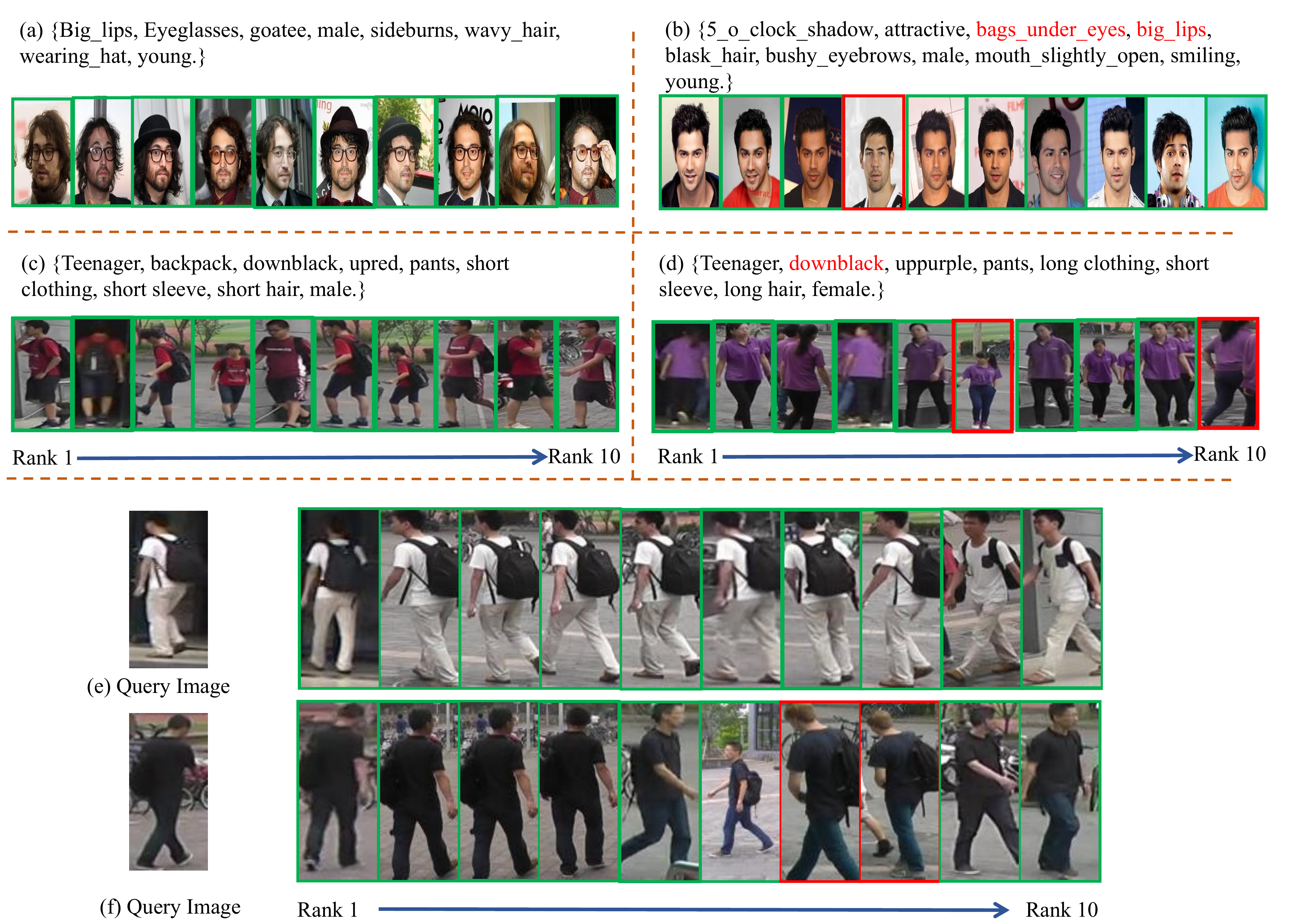}
	\end{center} %\linewidth
	\vskip -0.15in
	\caption{The examples of \textbf{A2I} on CelebA dataset and Market-1501 dataset, and \textbf{I2I} on Market-1501 dataset. True or false images are indicated by \textcolor{green}{green}/\textcolor{red}{red} boxes respectively. We highlight the attributes in \textcolor{red}{red} corresponding to the false matches.}
	\vskip -0.15in
	\label{fig:retrieval_results}
\end{figure*}

% This insures the attribute attention map can capture local attribute regions to improve the attribute-level discrimination of visual features.
Second, we show the examples of \textbf{A2I} on CelebA dataset and Market-1501 dataset, and \textbf{I2I} on Market-1501 dataset in Fig.~\ref{fig:retrieval_results}.
It can be seen that:
(i) Most of the retrieval images exactly match the attribute query in Fig.~\ref{fig:retrieval_results}(a) and Fig.~\ref{fig:retrieval_results}(b).
For example, MFHI accurately detects the local attributes (e.g., ``eyeglasses" and ``wavy\_hair") as presented the R@1 image in Fig.~\ref{fig:retrieval_results}(a).
Meanwhile, we also find a false retrieval result due to the unconspicuous local attributes.
For instance, it is hard to judge whether the R@4 image in Fig.~\ref{fig:retrieval_results}(b) possesses ``bags\_under\_eyes" attribute.
(ii) In addition, as seen from the retrieval results in Fig.~\ref{fig:retrieval_results}(d), due to the ambiguous visual appearance, there appear a few false retrieval results in re-ID.
For example, the R@6 and R@10 images in Fig.~\ref{fig:retrieval_results}(d) are ``downblue”, while the R@4 image in Fig.~\ref{fig:retrieval_results}(d) is ``downblack".
(iii) For the common image queries, our MFHI obtains more outstanding performances as shown in  Fig.~\ref{fig:retrieval_results}(e) and  Fig.~\ref{fig:retrieval_results}(f).
We can find that our MFHI correctly retrieves the target identity images in Fig.~\ref{fig:retrieval_results}(e), while the false matches R@7 and R@8 images in Fig.~\ref{fig:retrieval_results}(f) are due to very similar visual appearance of different identities. 

%------------------------------Table 5---------------------------------------------------------%
\begin{table*}[!htbp]
	\scriptsize
	\centering
	\vskip -0.2in
	\caption{Effect of local attribute-level discrimination (\%).}
		\vskip -0.05in
	\begin{tabular}{|c|c|c|c|c|c|c|c|c|c|c|c|c|c|}
		\hline
		Datasets & \multicolumn{6}{|c|}{CelebA} & \multicolumn{6}{|c|}{LFWA}   \\
		\hline
		\multirow{2}*{Methods} & \multicolumn{3}{|c|}{\textbf{Image$\rightarrow$Attribute}} & \multicolumn{3}{|c|}{\textbf{Attribute$\rightarrow$Image}} & \multicolumn{3}{|c|}{\textbf{Image$\rightarrow$Attribute}} & \multicolumn{3}{|c|}{\textbf{Attribute$\rightarrow$Image}} \\
		\cline{2-13}
		& Top-1 & Top-5 & Top-10 & R@1 & R@5 & R@10 & Top-1 & Top-5 & Top-10 & R@1 & R@5 & R@10 \\
		\hline
		GCL & 22.8 & 49.9 & 63.3 & 32.4 & 59.5 & 69.0 & 33.6 & 63.4 & 74.6 & 32.1 & 56.6 & 67.4 \\
		\hline
		MFHI  & \textbf{30.1} & \textbf{58.3} & \textbf{71.0} & \textbf{41.0} & \textbf{67.7} & \textbf{77.3} & \textbf{36.3} & \textbf{65.9} & \textbf{77.1} & \textbf{39.1} & \textbf{66.7} & \textbf{76.2}  
		\\
		\hline
	\end{tabular}

	\label{table:local_analysis}
\end{table*}
%------------------------------Table 5---------------------------------------------------------%
%------------------------------Table 5---------------------------------------------------------%
\begin{table*}[!htbp]

	\scriptsize
	\centering
	\vskip -0.15in
	\caption{Comparisons to the Self-Attention on person re-ID task(\%).}
		\vskip -0.05in
	\begin{tabular}{|c|c|c|c|c|c|c|c|c|c|c|c|c|c|c|c|}
		\hline
		Datasets & \multicolumn{7}{|c|}{Market-1501} & \multicolumn{7}{|c|}{DukeMTMC-reID}   \\
		\hline
		\multirow{2}*{Methods} & \multicolumn{3}{|c|}{\textbf{Image$\rightarrow$Attribute}} & \multicolumn{4}{|c|}{\textbf{Attribute$\rightarrow$Image}} & \multicolumn{3}{|c|}{\textbf{Image$\rightarrow$Attribute}} & \multicolumn{4}{|c|}{\textbf{Attribute$\rightarrow$Image}} \\
		\cline{2-15}
		& Top-1 & Top-5 & Top-10 & R@1 & R@5 & R@10 & mAP & Top-1 & Top-5 & Top-10 & R@1 & R@5 & R@10 & mAP\\
		\hline
		Self-Attention~\cite{zhang2019self} & 26.5 & 59.4 & 72.4 & 37.1 & 63.2 & 71.7 & 32.6 & 24.7 & 54.5 & 67.7 & 50.3 & 73.2 & 81.1 & 28.8 \\
		\hline
		MFHI  & \textbf{36.2} & \textbf{68.2} & \textbf{80.0} & \textbf{44.8} & \textbf{66.9} & \textbf{74.3} & \textbf{33.5} & \textbf{27.0} & \textbf{58.8} & \textbf{71.5} & \textbf{57.2} & \textbf{74.1} & \textbf{81.2} & \textbf{31.5}  
		\\
		\hline
	\end{tabular}

	\label{table:com_self_attention}
\end{table*}
%------------------------------Table 5---------------------------------------------------------%

%--------------------Figure 6---------------------------%
%--------------------Figure 6---------------------------%

\subsection{Ablation Study}
\textbf{Effectiveness of Semantics-Guided Spatial Attention.}
To evaluate the effectiveness of semantics-guided spatial attention in our MFHI, we examine two aspects of ablation study.
On one hand, we remove the spatial attention flow in MFHI, and then just extract the visual features with global category-level (i.e., GCL) discrimination.
The experimental results are reported in Table~\ref{table:local_analysis}.
It can be concluded that:
(i) Only global category-level discrimination is not sufficient to our tasks, although achieves higher performances than most compared methods.
(ii) The improvements obtained by MFHI to GCL (2.5\%$\sim$10.1\% on LFWA dataset and 7.3\%$\sim$8.6\% on CelebA dataset) are significant. 
Obviously, learning category-level discrimination and capturing discriminative visual regions jointly is the optimal solution for modality-free human identification problem.
Thus, local attribute-level discrimination with semantics-guided spatial attention is necessary.

On the other hand, to show the superiority of the proposed semantics-guided attention branch, we also replace the semantic attention in our MFHI by the popularly used self-attention in~\cite{zhang2019self} and conduct extensive experiments on person re-ID task. It can be seen from Tab.~\ref{table:com_self_attention} that our semantic-guided attention is better suited to such large scale identification tasks than self-attention.
The main reason lies in that the self-attention mechanism is usually applied to capture long-range dependencies and has been shown to be effective in generating high-resolution image details. However, for such large scale identification tasks, the main challenge is to maximize the separability of human identity. Hence, a semantics-guided spatial attention is enforced on visual modality in our MFHI to enhance local-level discrimination.

%--------------------Figure 7---------------------------%
\begin{figure}[!tbp]
		\vskip -0.2in
	\begin{center}
		\includegraphics[width=0.35\textwidth ]{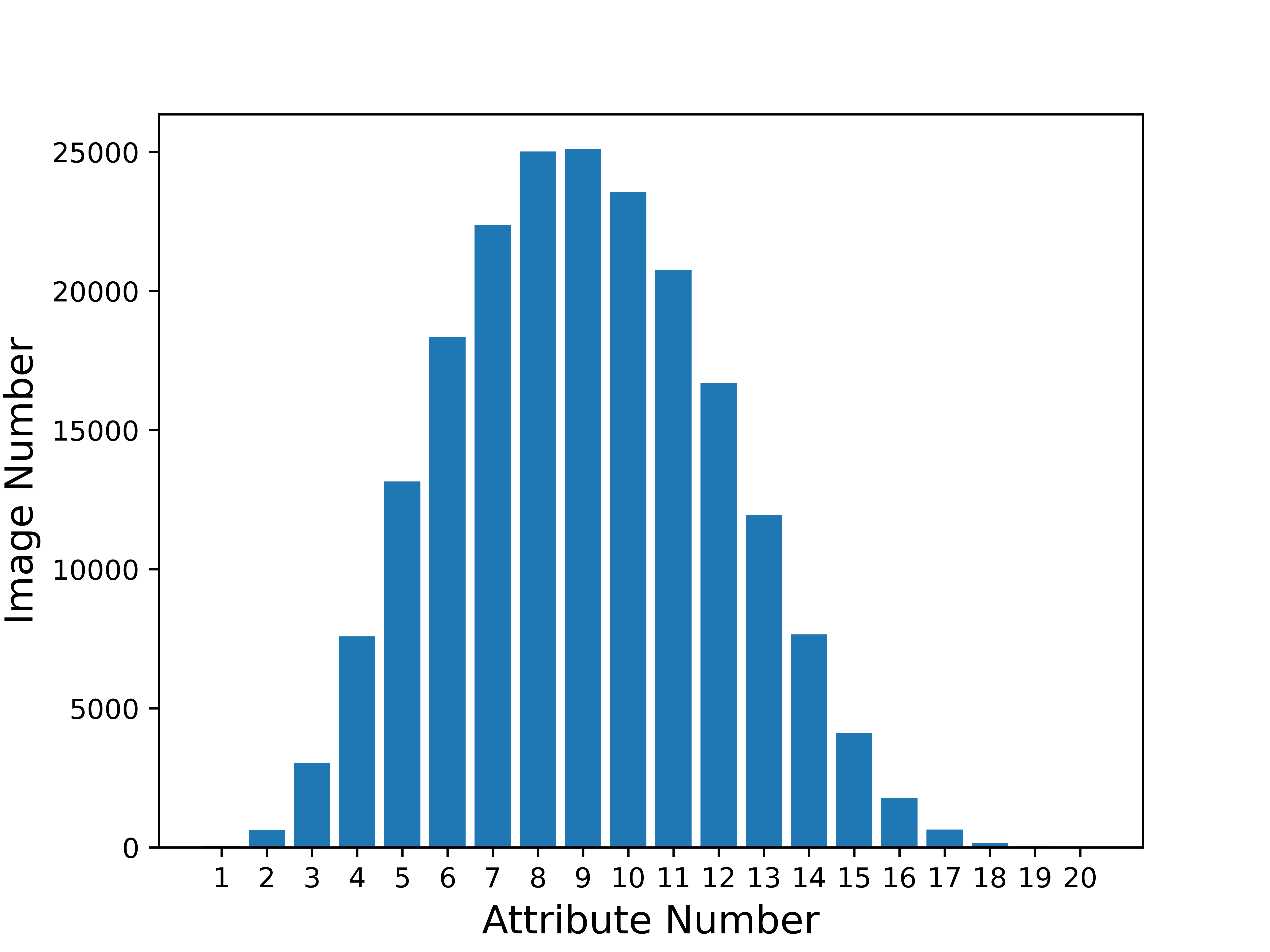}
	\end{center} %\linewidth
	\vskip -0.2in
	\caption{The statistics of attributes on CelebA dataset.}
	\label{fig:att_number}
\end{figure}
%--------------------Figure 7---------------------------%
\subsection{Parameter Analysis}

\textbf{Influence of the attribute activation map number (i.e., $D$).}
As observed from Fig.~\ref{fig:att_number}, the number of attributes in an image obeys normal distribution on CelebA dataset, where $\mu$ and $\sigma$ represent its mean and standard deviation respectively.
% Under \textbf{Image$\rightarrow$Attribute} and \textbf{Attribute$\rightarrow$Image} two setups,
To explore the influence of $D$ in our MFHI, we select a group of $D$ by referring to the above statistics, and then conduct \textbf{I2A} and \textbf{A2I} experiments on CelebA dataset, where $D \in \{ 5,10,14 \}$. 
The comparative results are reported in Table~\ref{tab:parameter_analysis}.
These results verify that:
(i) $D=10$ (located on $\left ( \mu - \sigma, \mu + \sigma \right )$) outperforms other settings with an obvious margin (0.10\%$\sim$1.13\%), except in R@10 of \textbf{A2I}.
% which play an important role in retrieving.
(ii) $D=5$ and $D=14$ achieve unsatisfied performances, since the smaller number of attribute activation maps cannot capture enough representative local regions, while the larger number often brings noises, thus learning indiscriminative features.
% When the number of attribute activation class maps considered is small, the discriminative attribute regions are not fully localized which decreases the local attribute-level discrimination of visual feature.
% In contrast, more attribute activation class maps may introducing some noisy information due to most images not possess so many attributes.

\begin{figure}[!tbp]
	\vskip -0.2in
	\begin{center}
		\includegraphics[width=0.4\textwidth ]{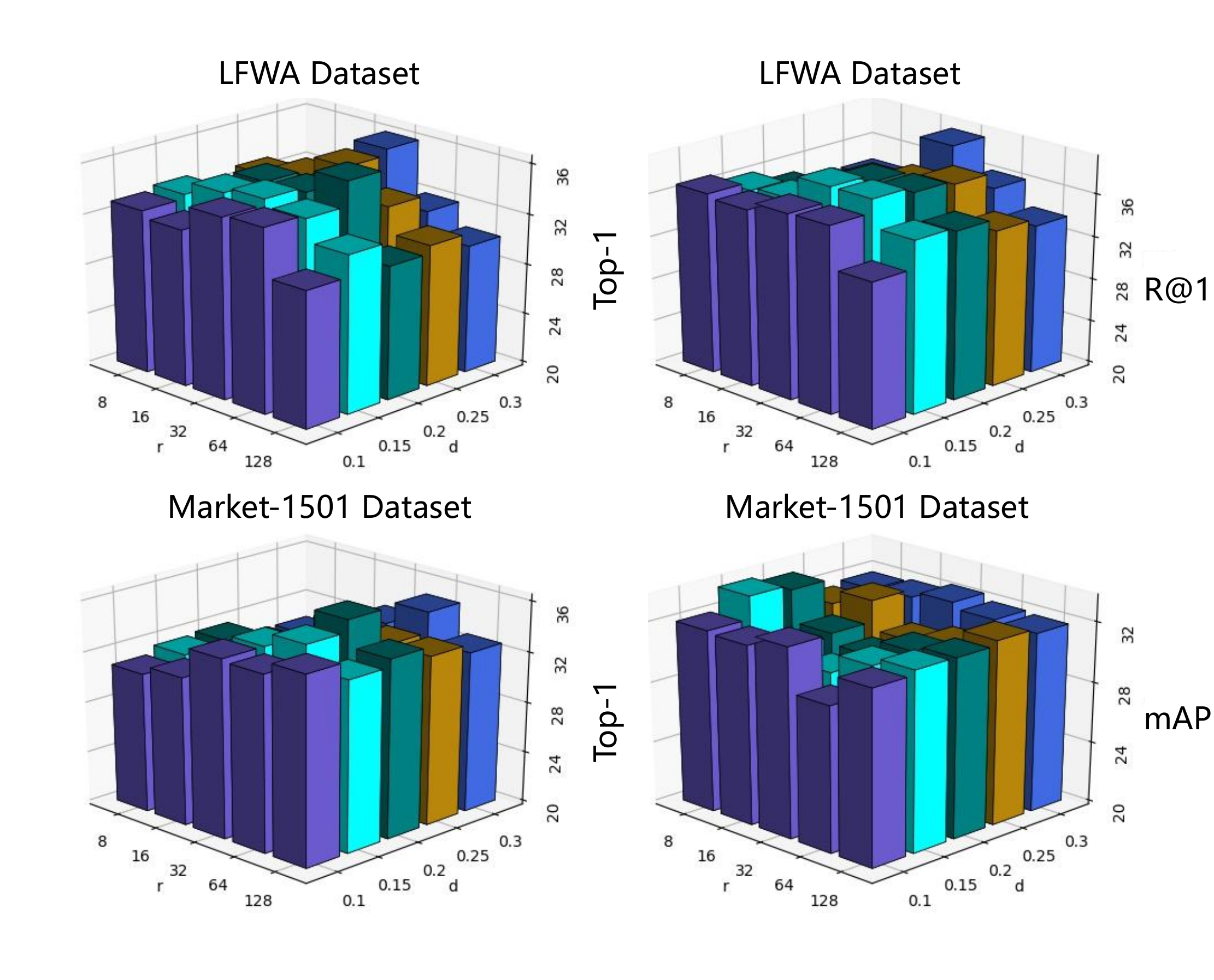}
	\end{center} %\linewidth	
	\vskip -0.15in
	\caption{The influence of $\left(r,d\right)$ on LFWA dataset and Market-1501 dataset.}
	\vskip -0.2in
	\label{fig:parameter_m_d}
\end{figure}

%------------------------------Table 6---------------------------------------------------------%
\begin{table}[!tbp]
	\footnotesize
	\centering
    \vskip -0.1in
	\caption{The influence of $D$ on our MFHI for CelebA dataset (\%).}
		\vskip -0.05in
	\begin{tabular}{|c|c|c|c|c|c|c|}
		\hline
		\multirow{2}*{Value of $D$}& \multicolumn{3}{|c|}{\textbf{Image$\rightarrow$Attribute}} & \multicolumn{3}{|c|}{\textbf{Attribute$\rightarrow$Image}} \\\cline{2-7}
		
		& Top-1 & Top-5 & Top-10 & R@1 & R@5 & R@10 \\
		\hline
		5 & 29.40 & 57.91 & 70.03 & 40.00 & 66.80 & \textbf{77.30} \\
		\hline
		10 & \textbf{30.09} & \textbf{58.25} & \textbf{70.98} & \textbf{41.00} & \textbf{67.70} & \textbf{77.30} \\
		\hline
		14 & 29.28 & 57.89 & 69.85 & 40.90 & 66.80 & 76.90\\
		\hline
	\end{tabular}
	\vskip -0.2in
	\label{tab:parameter_analysis}
\end{table}
%------------------------------Table 6---------------------------------------------------------%
\textbf{Influence of feature scale and margin (i.e., $r$ and $d$).}
For our MFHI, there still exist two parameters, i.e., $r$ and $d$ in Eq.~(\ref{dcm}). 
By varying $r$ from $\left \{ 8,16,32,64,128\right \}$, $d$ from $\left \{0.1,0.15,0.2,0.25,0.3\right \}$, and fixing other parameters as defaults, we run different models, and report both R@1 of \textbf{A2I} and Top-1 of \textbf{I2A} on LFWA dataset.
In addition, we also conduct the same experiments on Market-1501 dataset to analyze the parameter influence on Top-1 and mAP.
The experimental results are shown in Fig.~\ref{fig:parameter_m_d}, it can be observed that:
(i) $\left(64,0.2\right)$ and $\left(32,0.3\right)$ are the optimal parameter settings for the two challenging scenarios \textbf{I2A} and \textbf{A2I} on LFWA dataset respectively.
(ii) For Market-1501 dataset,  $\left(64,0.2\right)$ performs best in the \textbf{I2A} scenario, while $\left(8,0.15\right)$ is more suitable for the \textbf{A2I} scenario. 
The above discussions further demonstrate that by selecting appropriate parameters, our MFHI can be flexibly applied to perform different tasks.

%------------------------------Experiment---------------------------------------------------------%

\section{Conclusion}
We propose the first algorithmic framework for modality-free human identification task.
In particular, we take an initial attempt, and formulate it as a generic zero-shot learning model.
In addition, to maximize the human identity separability with interpretability, a semantics-guided spatial attention mechanism is enforced on visual modality.
% Furthermore, we bridge the visual image and the textual description by learning a shared space, and then maximize the distribution consistency between different modalities via learned identity prototypes.
Furthermore, we bridge different modalities by learning a shared space, and then maximize their distribution consistency via learned identity prototypes.
We have conducted extensive experiments on two challenging identification tasks, (i.e., face identification and re-ID), and achieve promising results compared with other alternatives.
In essence, learning one prototype for an identity is generally insufficient to recognize one identity and differentiate two identities.
Thus, our ongoing research work includes learning prototypes adaptively with the data distribution.
%--------------------Figure 8---------------------------%
%--------------------Figure 8---------------------------%
% or


%\appendices

%\section{}



\ifCLASSOPTIONcaptionsoff
  \newpage
\fi

\bibliographystyle{IEEEtran}
\bibliography{mybib}

\begin{thebibliography}{10}
\providecommand{\url}[1]{#1}
\csname url@samestyle\endcsname
\providecommand{\newblock}{\relax}
\providecommand{\bibinfo}[2]{#2}
\providecommand{\BIBentrySTDinterwordspacing}{\spaceskip=0pt\relax}
\providecommand{\BIBentryALTinterwordstretchfactor}{4}
\providecommand{\BIBentryALTinterwordspacing}{\spaceskip=\fontdimen2\font plus
\BIBentryALTinterwordstretchfactor\fontdimen3\font minus
  \fontdimen4\font\relax}
\providecommand{\BIBforeignlanguage}[2]{{%
\expandafter\ifx\csname l@#1\endcsname\relax
\typeout{** WARNING: IEEEtran.bst: No hyphenation pattern has been}%
\typeout{** loaded for the language `#1'. Using the pattern for}%
\typeout{** the default language instead.}%
\else
\language=\csname l@#1\endcsname
\fi
#2}}
\providecommand{\BIBdecl}{\relax}
\BIBdecl

\bibitem{deng2019arcface}
J.~Deng, J.~Guo, N.~Xue, and S.~Zafeiriou, ``Arcface: Additive angular margin
  loss for deep face recognition,'' in \emph{Proc. CVPR}, 2019.

\bibitem{6737218}
Z.~{Lai}, Y.~{Xu}, Z.~{Jin}, and D.~{Zhang}, ``Human gait recognition via
  sparse discriminant projection learning,'' \emph{IEEE Transactions on
  Circuits and Systems for Video Technology}, vol.~24, no.~10, pp. 1651--1662,
  2014.

\bibitem{zeng2020hierarchical}
K.~Zeng, M.~Ning, Y.~Wang, and Y.~Guo, ``Hierarchical clustering with
  hard-batch triplet loss for person re-identification,'' in \emph{Proc. CVPR},
  2020.

\bibitem{quach2021dyglip}
K.~G. Quach, P.~Nguyen, H.~Le, T.-D. Truong, C.~N. Duong, M.-T. Tran, and
  K.~Luu, ``Dyglip: A dynamic graph model with link prediction for accurate
  multi-camera multiple object tracking,'' in \emph{Proc. CVPR}, 2021.

\bibitem{Chen2010Time}
C.~C. Loy, T.~Xiang, and S.~Gong, ``Time-delayed correlation analysis for
  multi-camera activity understanding,'' \emph{International Journal of
  Computer Vision}, vol.~90, no.~1, pp. 106--129, 2010.

\bibitem{zhang2017learning}
L.~Zhang, T.~Xiang, and S.~Gong, ``Learning a deep embedding model for
  zero-shot learning,'' in \emph{Proc. CVPR}, 2017.

\bibitem{Xie_2019_CVPR}
G.-S. Xie, L.~Liu, and el~al, ``Attentive region embedding network for
  zero-shot learning,'' in \emph{Proc. CVPR}, 2019.

\bibitem{annadani2018preserving}
Y.~Annadani and S.~Biswas, ``Preserving semantic relations for zero-shot
  learning,'' in \emph{Proc. CVPR}, 2018.

\bibitem{sung2018learning}
F.~Sung, Y.~Yang, L.~Zhang, and et~al, ``Learning to compare: Relation network
  for few-shot learning,'' in \emph{Proc. CVPR}, 2018.

\bibitem{8692748}
H.~Guo, K.~Zhu, M.~Tang, and J.~Wang, ``Two-level attention network with
  multi-grain ranking loss for vehicle re-identification,'' \emph{IEEE
  Transactions on Image Processing}, vol.~28, no.~9, pp. 4328--4338, 2019.

\bibitem{cao2018vggface2}
Q.~Cao, L.~Shen, W.~Xie, O.~M. Parkhi, and A.~Zisserman, ``Vggface2: A dataset
  for recognising faces across pose and age,'' in \emph{Proc. FG}, 2018.

\bibitem{schroff2015facenet}
F.~Schroff, D.~Kalenichenko, and J.~Philbin, ``Facenet: A unified embedding for
  face recognition and clustering,'' in \emph{Proc. CVPR}, 2015.

\bibitem{xu2021consistent}
X.~Xu, Y.~Huang, P.~Shen, and et~al, ``Consistent instance false positive
  improves fairness in face recognition,'' in \emph{Proc. CVPR}, 2021.

\bibitem{lin2019improving}
Y.~Lin, L.~Zheng, Z.~Zheng, Y.~Wu, Z.~Hu, C.~Yan, and Y.~Yang, ``Improving
  person re-identification by attribute and identity learning,'' \emph{Pattern
  Recognition}, vol.~95, pp. 151--161, 2019.

\bibitem{10.1145/3383184}
Z.~Zheng, L.~Zheng, M.~Garrett, Y.~Yang, M.~Xu, and Y.-D. Shen, ``Dual-path
  convolutional image-text embeddings with instance loss,'' \emph{ACM
  Transactions on Multimedia Computing, Communications, and Applications},
  vol.~16, no.~2, pp. 1--23, 2020.

\bibitem{chai2021video}
T.~Chai, Z.~Chen, A.~Li, J.~Chen, X.~Mei, and Y.~Wang, ``Video person
  re-identification using attribute-enhanced features,'' \emph{arXiv preprint
  arXiv:2108.06946}, 2021.

\bibitem{dong2019person}
Q.~Dong, S.~Gong, and X.~Zhu, ``Person search by text attribute query as
  zero-shot learning,'' in \emph{Proc. ICCV}, 2019.

\bibitem{8931651}
X.~Zhang, S.~Gui, Z.~Zhu, Y.~Zhao, and J.~Liu, ``Hierarchical prototype
  learning for zero-shot recognition,'' \emph{IEEE Transactions on Multimedia},
  vol.~22, no.~7, pp. 1692--1703, 2020.

\bibitem{LIU2020103924}
Z.~Liu, X.~Zhang, Z.~Zhu, S.~Zheng, Y.~Zhao, and J.~Cheng, ``Convolutional
  prototype learning for zero-shot recognition,'' \emph{Image and Vision
  Computing}, vol.~98, p. 103924, 2020.

\bibitem{chen2021free}
S.~Chen, W.~Wang, B.~Xia, and et~al, ``Free: Feature refinement for generalized
  zero-shot learning,'' in \emph{Proc. ICCV}, 2021.

\bibitem{chen2021hsva}
S.~Chen, G.-S. Xie, Y.~Liu, Q.~Peng, B.~Sun, H.~Li, X.~You, and L.~Shao,
  ``Hsva: Hierarchical semantic-visual adaptation for zero-shot learning,''
  \emph{arXiv preprint arXiv:2109.15163}, 2021.

\bibitem{tay2019aanet}
C.-P. Tay, S.~Roy, and K.-H. Yap, ``Aanet: Attribute attention network for
  person re-identifications,'' in \emph{Proc. CVPR}, 2019.

\bibitem{yang2021sega}
F.~Yang, R.~Wang, and X.~Chen, ``Sega: Semantic guided attention on visual
  prototype for few-shot learning,'' \emph{arXiv preprint arXiv:2111.04316},
  2021.

\bibitem{ge2021semantic}
J.~Ge, H.~Xie, S.~Min, and Y.~Zhang, ``Semantic-guided reinforced region
  embedding for generalized zero-shot learning,'' in \emph{Proc. AAAI}, 2021.

\bibitem{zhou2016learning}
B.~Zhou, A.~Khosla, A.~Lapedriza, A.~Oliva, and A.~Torralba, ``Learning deep
  features for discriminative localization,'' in \emph{Proc. CVPR}, 2016.

\bibitem{faghri2017vse++}
F.~Faghri, D.~J. Fleet, J.~R. Kiros, and S.~Fidler, ``Vse++: Improving
  visual-semantic embeddings with hard negatives,'' in \emph{Proc. BMVC}, 2018.

\bibitem{li2019visual}
K.~Li, Y.~Zhang, K.~Li, Y.~Li, and Y.~Fu, ``Visual semantic reasoning for
  image-text matching,'' in \emph{Proc. ICCV}, 2019.

\bibitem{lee2018stacked}
K.-H. Lee, X.~Chen, G.~Hua, H.~Hu, and X.~He, ``Stacked cross attention for
  image-text matching,'' in \emph{Proc. ECCV}, 2018.

\bibitem{liu2015deep}
Z.~Liu, P.~Luo, X.~Wang, and X.~Tang, ``Deep learning face attributes in the
  wild,'' in \emph{Proc. ICCV}, 2015.

\bibitem{LFWTech}
G.~B. Huang, M.~Ramesh, T.~Berg, and E.~Learned-Miller, ``Labeled faces in the
  wild: A database for studying face recognition in unconstrained
  environments,'' Tech. Rep. 07-49, 2007.

\bibitem{hu2018squeeze}
J.~Hu, L.~Shen, and G.~Sun, ``Squeeze-and-excitation networks,'' in \emph{Proc.
  CVPR}, 2018.

\bibitem{8578734}
P.~{Anderson}, X.~{He}, C.~{Buehler}, D.~{Teney}, M.~{Johnson}, S.~{Gould}, and
  L.~{Zhang}, ``Bottom-up and top-down attention for image captioning and
  visual question answering,'' in \emph{Proc. CVPR}, 2018.

\bibitem{zheng2015scalable}
L.~Zheng, L.~Shen, L.~Tian, S.~Wang, J.~Wang, and Q.~Tian, ``Scalable person
  re-identification: A benchmark,'' in \emph{Proc. ICCV}, 2015.

\bibitem{ristani2016performance}
E.~Ristani, F.~Solera, R.~Zou, and el~al, ``Performance measures and a data set
  for multi-target, multi-camera tracking,'' in \emph{Proc. ECCV}, 2016.

\bibitem{andrew2013deep}
G.~Andrew, R.~Arora, J.~Bilmes, and K.~Livescu, ``Deep canonical correlation
  analysis,'' in \emph{Proc. ICML}, 2013.

\bibitem{eisenschtat2017linking}
A.~Eisenschtat and L.~Wolf, ``Linking image and text with 2-way nets,'' in
  \emph{Proc. CVPR}, 2017.

\bibitem{tolstikhin2016minimax}
I.~O. Tolstikhin, B.~K. Sriperumbudur, and et~al, ``Minimax estimation of
  maximum mean discrepancy with radial kernels,'' in \emph{Proc. NIPS}, 2016.

\bibitem{sun2016deep}
B.~Sun and K.~Saenko, ``Deep coral: Correlation alignment for deep domain
  adaptation,'' in \emph{Proc. ECCV}, 2016.

\bibitem{ijcai2018-153}
Z.~Yin, W.-S. Zheng, A.~Wu, and et~al, ``Adversarial attribute-image person
  re-identification,'' in \emph{Proc. IJCAI}, 2018.

\bibitem{li2017person}
S.~Li, T.~Xiao, H.~Li, B.~Zhou, D.~Yue, and X.~Wang, ``Person search with
  natural language description,'' in \emph{Proc. CVPR}, 2017.

\bibitem{li2017identity}
S.~Li, T.~Xiao, H.~Li, W.~Yang, and X.~Wang, ``Identity-aware textual-visual
  matching with latent co-attention,'' in \emph{Proc. ICCV}, 2017.

\bibitem{Zhong2017RerankingPR}
Z.~Zhong, L.~Zheng, D.~Cao, and S.~Li, ``Re-ranking person re-identification
  with k-reciprocal encoding,'' in \emph{Proc. CVPR}, 2017.

\bibitem{Su_2017_ICCV}
C.~Su, J.~Li, S.~Zhang, J.~Xing, W.~Gao, and Q.~Tian, ``Pose-driven deep
  convolutional model for person re-identification,'' in \emph{Proc. ICCV},
  2017.

\bibitem{8265515}
Y.~{Chen}, X.~{Zhu}, and S.~{Gong}, ``Person re-identification by deep learning
  multi-scale representations,'' in \emph{Proc. ICCVW}, 2017.

\bibitem{8466034}
L.~{Wei}, S.~{Zhang}, H.~{Yao}, W.~{Gao}, and Q.~{Tian}, ``Glad:
  Global–local-alignment descriptor for scalable person re-identification,''
  \emph{IEEE Transactions on Multimedia}, vol.~21, no.~4, pp. 986--999, 2019.

\bibitem{8578818}
Y.~{Shen}, T.~{Xiao}, H.~{Li}, and et~al, ``End-to-end deep kronecker-product
  matching for person re-identification,'' in \emph{Proc. CVPR}, 2018.

\bibitem{Sun2018BeyondPM}
Y.~Sun, L.~Zheng, Y.~Yang, Q.~Tian, and S.~Wang, ``Beyond part models: Person
  retrieval with refined part pooling,'' in \emph{Proc. ECCV}, 2018.

\bibitem{Almazn2018ReIDDR}
J.~Almaz{\'a}n, B.~Gajic, N.~Murray, and D.~Larlus, ``Re-id done right: towards
  good practices for person re-identification,'' \emph{arXiv preprint
  arXiv:1801.05339}, 2018.

\bibitem{Kalayeh2018HumanSP}
M.~M. Kalayeh, E.~Basaran, M.~Gokmen, and et~al, ``Human semantic parsing for
  person re-identification,'' in \emph{Proc. CVPR}, 2018.

\bibitem{zhong2017random}
Z.~Zhong, L.~Zheng, G.~Kang, S.~Li, and Y.~Yang, ``Random erasing data
  augmentation,'' \emph{arXiv preprint arXiv:1708.04896}, 2017.

\bibitem{zhang2019self}
H.~Zhang, I.~Goodfellow, D.~Metaxas, and A.~Odena, ``Self-attention generative
  adversarial networks,'' in \emph{Proc. ICML}, 2019.

\end{thebibliography}
\end{document}